\documentclass{article}
\usepackage[final]{neurips_2025}
\usepackage{amsmath}

\usepackage{amsmath,amsfonts,bm}

\def\eqref#1{equation~\ref{#1}}

\def\1{\bm{1}}

\DeclareMathAlphabet{\mathsfit}{\encodingdefault}{\sfdefault}{m}{sl}
\SetMathAlphabet{\mathsfit}{bold}{\encodingdefault}{\sfdefault}{bx}{n}

\DeclareMathOperator*{\argmax}{arg\,max}

\usepackage[utf8]{inputenc} 
\usepackage[T1]{fontenc}    
\usepackage{tgtermes}
\usepackage{wrapfig}
\usepackage{tcolorbox}
\usepackage{booktabs}       
\usepackage{amsfonts}       
\usepackage{nicefrac}       
\usepackage{subcaption}
\usepackage{microtype}      
\usepackage{tabularray}
\usepackage{enumerate}
\usepackage{enumitem}
\setlist{leftmargin=1ex}
\usepackage{mathtools}
\usepackage{amsthm, amssymb}

\usepackage{algorithm} 
\usepackage[noend]{algpseudocode}
\usepackage{xspace}
\usepackage{multirow}
\usepackage{lipsum}

\usepackage[normalem]{ulem}
\usepackage{mdframed}
\usepackage{xcolor}
\definecolor{babyblueeyes}{rgb}{0.63, 0.79, 0.95}
\definecolor{brightpink}{HTML}{D8315B}
\definecolor{lightpink}{HTML}{EF798A}
\definecolor{cvprblue}{rgb}{0.21,0.49,0.74}
\definecolor{myblue}{HTML}{d8ebf8}
\definecolor{lightred}{HTML}{D33E43}
\definecolor{mygreen}{HTML}{2DD881}
\definecolor{darkgreen}{HTML}{006400}
\definecolor{salmon}{HTML}{FA8072}
\definecolor{mybluee}{RGB}{0, 102, 204}
\definecolor{citeblue}{HTML}{002A88}
  
\usepackage[pagebackref,breaklinks,colorlinks,allcolors=citeblue]{hyperref}
\usepackage{graphicx}
\usepackage{url}

\usepackage{caption}
\usepackage{etoolbox}
\newtoggle{showlabels}
\toggletrue{showlabels} 

 \title{Does Thinking More \emph{Always} Help? \\ Mirage of Test-Time Scaling in Reasoning Models}

\author{
  Soumya Suvra Ghosal$^1$\thanks{Equal Contribution.}~, Souradip Chakraborty$^{1*}$, Avinash Reddy$^5$, Yifu Li$^3$, \\ \textbf{Mengdi Wang$^3$, Dinesh Manocha$^1$, Furong Huang$^{1, 3}$,}
\\ \textbf{Mohammad Ghavamzadeh$^4$, Amrit Singh Bedi$^5$} \\
$^1$University of Maryland, $^2$Princeton University, \\ $^3$Capital One, $^4$Amazon AGI, $^5$University of Central Florida \\
}

\begin{document}

\maketitle

\begin{abstract}

Recent trends in test-time scaling for reasoning models (e.g., OpenAI o1, DeepSeek R1) have led to a popular belief that extending thinking traces using prompts like “Wait” or “Let me rethink” can improve performance.  This raises a natural question: \textit{Does thinking more at test-time truly lead to better reasoning?} To answer this question, we perform a detailed empirical study across models and benchmarks, which reveals a consistent pattern of initial performance improvements from additional thinking followed by a decline, due to `\textit{overthinking}’. To understand this non-monotonic trend, we consider a simple probabilistic model, which reveals that additional thinking increases output variance, creating an illusion of improved reasoning while ultimately undermining precision. Thus, observed gains from ``more thinking'' are not true indicators of improved reasoning, causing a \textit{mirage effect}, but artifacts stemming from the connection between model uncertainty and evaluation metric. This suggests that test-time scaling through extended thinking is not an effective way to utilize the inference thinking budget. Recognizing these limitations, we introduce an alternative test-time scaling approach, \textit{parallel thinking}, inspired by Best-of-N sampling. Our method generates multiple independent reasoning paths within the same inference budget and selects the most consistent response via majority vote, achieving up to $20\%$ higher accuracy compared to extended thinking. This provides a simple yet effective mechanism for test-time scaling of reasoning models.
\end{abstract}

\section{Test-time Scaling Capability in LLMs for Reasoning}
\label{sec:introduction}

\begin{wrapfigure}{r}{0.45\textwidth}
\vspace{-7mm}
  \begin{center}
    \includegraphics[width=0.424\textwidth]{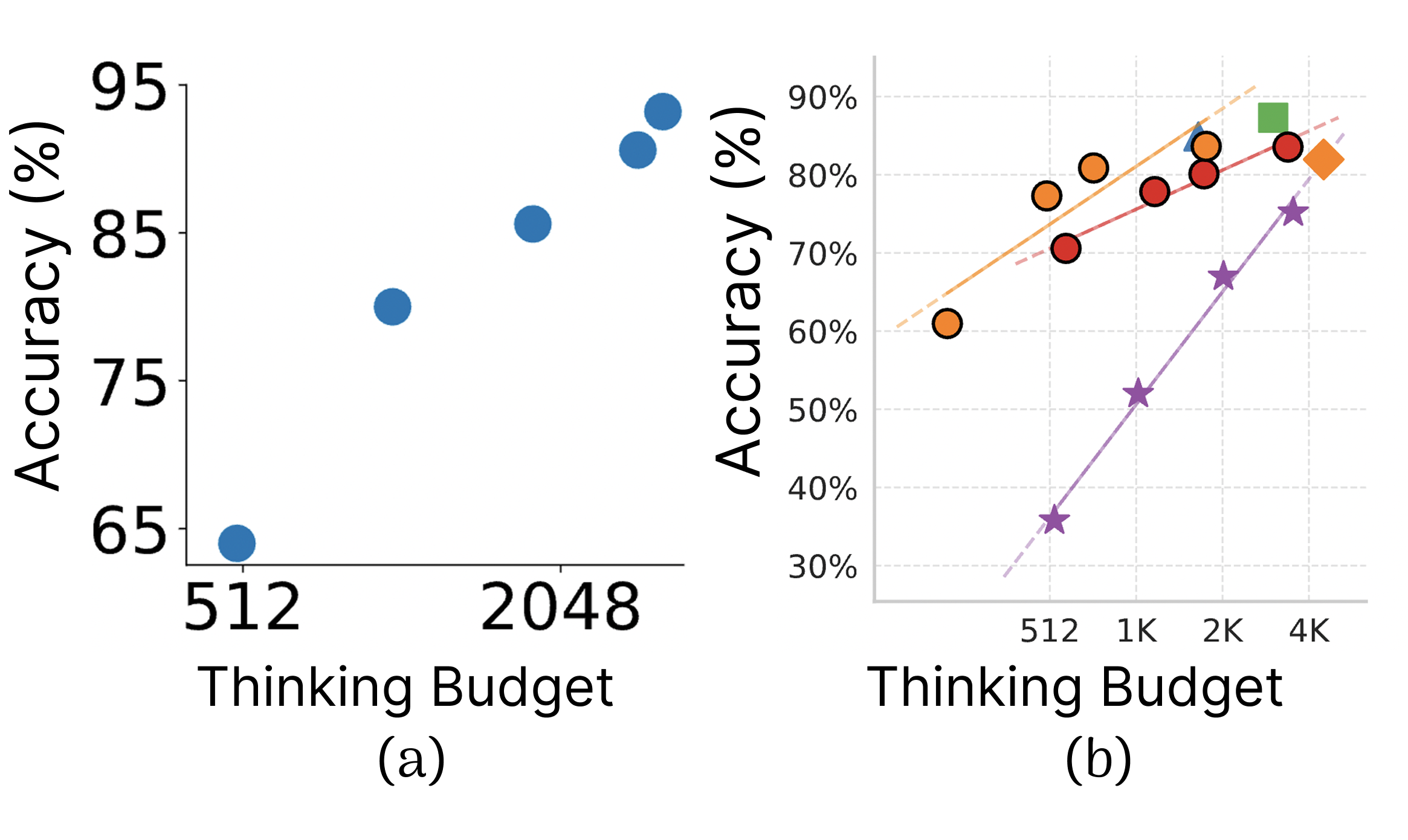}\vspace{-3mm}
  \end{center}
  \caption{(a) Scaling behavior observed in \citet{muennighoff2025s1}. (b) Trends in \citet{aggarwal2025l1} (on MATH-500~\citep{hendrycks2021measuring}).}
  \label{fig:first}
  \vspace{-4mm}
\end{wrapfigure} Recent improvements in the performance of large language models (LLMs) on reasoning tasks have been primarily driven by enhancements in their reasoning capabilities through reinforcement learning-based training ~\citep{guo2025deepseek, shao2024deepseekmath}. With increasingly capable reasoning models, a key question that arises is how to effectively scale their performance at test time. Recent literature \citep{muennighoff2025s1} has shown that extending the thinking (or reasoning) process at test-time through explicit prompts such as \textit{``Wait''}, \textit{``Let me rethink''}, or \textit{``Think more''} significantly improves performance. This approach, requiring no additional supervision, has led to the prevailing belief that more thinking at test-time inherently improves accuracy (see Figure~\ref{fig:first}). 

\textbf{Incomplete picture of test-time scaling}. While appealing, the above narrative in prior works presents an incomplete picture. In contrast to prior claims, our empirical investigation uncovers a nuanced phenomenon: extending thinking at test-time initially boosts model accuracy, but performance degrades subsequently with prolonged thinking (cf. Section \ref{subsec:observations}). This non-monotonic behavior (a clear pattern consistent across various tasks and datasets (cf. Figure~\ref{fig:wait_think_more}) reveals the presence of a \textit{critical point} in the length of the thinking trace beyond which performance declines, which we call `overthinking', largely unrecognized by existing research. These observations raise a fundamental question: \textit{Why does additional thinking beyond a certain point degrade the model performance?}

\textbf{Understanding overthinking: a variance-based explanation.} 
To answer the above question, we take a step back and analyze a simple one-dimensional probabilistic framework (cf. Section~\ref{sec:mirage}), examining how changes in the variance of the sampling distribution affect the expected value of a target reward. Interestingly, we observe that as the variance increases from low to high, the expected reward exhibits a similar non-monotonic pattern: initially increasing, then decreasing (cf. Figure \ref{fig:mirage_scaling}). Inspired by this insight, we empirically assess the variance of reasoning-model outputs under extended thinking by measuring the entropy of their output distributions. Our results clearly demonstrate that extended thinking significantly increases the variance of response distribution (cf. Figure \ref{fig:entropy}). This explains why average accuracy first improves and then deteriorates, revealing that the apparent gains from extended thinking reflect an illusion rather than genuine improvements in reasoning capability.

\textbf{Overthinking is inefficient for test-time scaling under a fixed budget.} These insights reveal a deeper inefficiency: extending a single reasoning trace is not an optimal use of the test-time compute budget. Because performance does not improve monotonically with more tokens, there is no reliable stopping criterion, making this strategy brittle in practice. 

\textbf{A fix: parallel thinking as a principled alternative.} To overcome these limitations, we propose \textit{parallel thinking}, a test-time scaling strategy inspired by Best-of-$N$ sampling~\citep{beirami2024theoretical, amini2024variational, nakano2021webgpt, stiennon2020learning, bon3, bon4}. Instead of continuing one thinking trace, we allocate the same token budget across multiple independent thinking paths and select the final answer via majority voting. This approach avoids entropy overgrowth, mitigates the overthinking trap, and achieves significantly better performance. For example, under a 16K token budget, parallel thinking yields up to 22\% higher accuracy compared to sequential scaling (Figure~\ref{fig:parallel_thinking}).
We summarize our contributions as follows.

    \textbf{(i) Empirical diagnosis of overthinking:} We investigate test-time scaling by encouraging extended thinking in state-of-the-art reasoning models with prompts as ``Wait'', ``Think more'' etc. It reveals a consistent non-monotonic trend in performance across multiple tasks and datasets (cf. Section \ref{sec:results}).
    
 \textbf{(ii) Illusion of test-time scaling:  alternative explanation}: We provide an interesting explanation for the non-monotonic trend of test-time scaling in reasoning models through a simple probabilistic framework. Our analysis clarifies why extending reasoning initially improves performance but eventually leads to degradation, highlighting variance as the key driver of the observed non-monotonic behavior (cf.  Section \ref{sec:mirage}).

     \textbf{ (iii) Variance-driven explanation of performance degradation in reasoning models:} By analyzing the entropy of the response distribution generated by the reasoning models, supported by our insights,  we show that extended thinking increases the variance of the model's output distribution. While this variance increase initially aligns with improved performance, it eventually disrupts reward alignment, explaining the degradation observed beyond a certain point (cf. Section \ref{variace_role}).

 \textbf{ (iv) Effective budget-control via parallel thinking:} We propose an alternative test-time scaling strategy, \textit{parallel thinking}, inspired by Best-of-$N$ sampling. By simultaneously generating multiple independent reasoning paths, this approach circumvents the pitfalls of sequential overthinking and yields higher performance, demonstrating genuine self-improvement capabilities (cf. Section~\ref{sec:effective_test}). This approach outperforms overthinking across all benchmarks and provides a reliable mechanism for inference-time scaling (Figure~\ref{fig:parallel_thinking}).

\section{Analysis of Test-Time Scaling in SoTA Reasoning LLMs}
\label{sec:results}
\textbf{Mathematical formulation of thinking process}: We begin with a mathematical representation of the thinking process generated by reasoning models, specified as $x \rightarrow z \rightarrow y,$
where $x$ is the input prompt, $z \sim \pi(\cdot|x)$ is the intermediate thinking trace, $y \sim \pi(\cdot|x, z)$ is the final answer, and $\pi$ denotes the language model. The RL objective for training reasoning models is:
\begin{align}
    \max_{\theta} \mathbb{E}_{x, z \sim \pi_\theta(\cdot|x), y \sim \pi_\theta(\cdot|x, z)} [R(x,y)]   ,
\end{align}
where $\pi_{\theta}$ is the parameterized model and $R(x, y)$ represents the true reward function (for example, an indicator to check if $y$ is correct or not), which is obtained once the policy generates the final response $y$. In practice~\citep{guo2025deepseek},  an additional format reward term, $R_{\text{format}}(x, y)$, is also often incorporated to encourage the model to generate outputs in a structured format: $x \rightarrow z \rightarrow y$. That is to make sure that the model first produces a ``thinking'' trace $z$, followed by a special end-of-thinking token (\texttt{</think>}), and then the final answer $y$. This process of first generating a thinking trace and then answering $x \rightarrow z \rightarrow y$ is referred to as ``standard thinking'' in this work.%

\textbf{Test time scaling.}  The test-time scaling extends the thinking process as 
\begin{align}
    x \rightarrow z_1 \rightarrow z_2 \rightarrow \cdots \rightarrow z_k \rightarrow y,
\end{align}
where, given the prompt $x$, the first thinking step is generated as $z_1 \sim \pi_\theta(\cdot|x)$. Then, instead of directly producing the final answer, the model is instructed to continue thinking using special tokens such as \textit{``Wait''} or \textit{``Think more''}, denoted as $\text{p}$ with the next thinking step sampled as $z_2 \sim \pi_\theta(\cdot|x, z_1, p)$. The process is continued iteratively for $k$ steps, and the model generates the final response as $y \sim \pi_\theta(\cdot|x, z_1, p, z_2, p, \ldots, z_k)$. For simplicity of representation, we just keep the thinking traces in the condition as $y \sim \pi_\theta(\cdot|x, z_1, z_2, \ldots, z_k)$, omitting the special instruction tokens $p$.

As discussed in the introduction, to gain a complete understanding of test-time scaling in reasoning language models, we begin with a systematic analysis of state-of-the-art reasoning models~\citep{guo2025deepseek}. This analysis requires two core components: the choice of reasoning models and the strategies used to control the test-time compute budget. We first introduce the budget control techniques we evaluate, then describe the reasoning models under study, and finally highlight key takeaways and novel insights that emerge from this investigation.

\subsection{Test-Time Budget Control (TTBC)} \label{budget_control_detail}
 To systematically probe how test-time budgets modulate model behavior, we apply two different budget control approaches on the model’s \emph{thinking} tokens~\citep{muennighoff2025s1} detailed as follows. 
\begin{itemize}
\item \textbf{TTBC 1: Wait \& Think more.} In this approach, we do not impose any explicit budget constraint on the number of thinking tokens, apart from the model's inherent maximum token limit ($32K$). Specifically for this approach, whenever the model attempts to generate the end-of-thinking delimiter (\texttt{</think>}), we suppress it, append the token \textit{``Wait''} to the thinking trace, and feed the modified trace back to the model. This intervention is applied iteratively, encouraging the model to extend its thinking. In this setup, there is no explicit budget for the thinking tokens, except the max token limit of the model, and we only modify the number of times \textit{``Wait''} is appended to the thinking trace.

\item \textbf{TTBC 2: Exact thinking tokens.} We enforce an exact thinking token budget of \(t_{\text{exact}}\) for each reasoning trajectory. Specifically, we iteratively append the token \textit{``Wait''} to the reasoning trace until the cumulative count of thinking tokens reaches exactly \(t_{\text{exact}}\). Once this threshold is reached, we terminate the thinking stage by allowing the end-of-thinking delimiter (\texttt{</think>}) to pass, signaling the model to generate the final response. This approach ensures that every reasoning trajectory is precisely constrained to \(t_{\text{exact}}\) tokens. For this setup, we vary \(t_{\text{exact}}\) across \([256, 512, 1024, 2048, 4096, 8192, 16384]\).

\end{itemize}

\begin{figure}[!ht]
    \centering
    \includegraphics[width=\linewidth]{figures/wait_think_more_scaling.pdf}
    \caption{\small\textbf{Results for \emph{Wait \& Think more.}} To investigate the variation in accuracy with increasing average thinking tokens generated during inference, we plot accuracy as a function of thinking tokens across three datasets: GSM-8K~\citep{cobbe2021gsm8k}, MATH-500~\citep{lightman2023let}, and AIME~\citep{AIME}. The average has been taken over all prompts in the test dataset. We report results for three SoTA open-source reasoning models: DeepSeek-R1-Distill-Qwen-1.5B (1st row),  DeepSeek-R1-Distill-Qwen-7B (2nd row), and DeepSeek-R1-Distill-Llama-8B model (3rd row)~\citep{guo2025deepseek}. During test-time, following \citet{muennighoff2025s1}, we induce the model to generate more thinking tokens by suppressing the end-of-thinking token delimiter (\texttt{</think>}) and instead appending \textit{``Wait''} to the model’s current thinking trace to encourage additional thinking. The baseline with no budget control (standard thinking) is marked by {{$\star$}}. Successive \textbf{\textcolor{salmon}{salmon circles}}  to the right of the star correspond to one, two, \dots\ additional \textit{``Wait''} insertions to the model's thinking trace. We note that to obtain the circles on the left of the {{$\star$}}, we enforce budgeting via controlling the maximum number of thinking tokens. From our experiments, we note that accuracy initially rises as the model is prompted to think longer, but then gradually falls off for all settings.}
\label{fig:wait_think_more}
\end{figure}

\begin{figure}[!ht]
    \centering
    \includegraphics[width=\linewidth]{figures/exact_thinking.pdf}
    \caption{\textbf{Results for \emph{Exact thinking tokens} setup.} We visualize accuracy as a function of the thinking token budget for three benchmark datasets: GSM-8K~\citep{cobbe2021gsm8k}, MATH-500~\citep{lightman2023let}, and AIME~\citep{AIME}. We report results for two open-source reasoning models: DeepSeek-R1-Distill-Qwen-1.5B (1st row), DeepSeek-R1-Distill-Qwen-7B (2nd row), and  DeepSeek-R1-Distill-Llama-8B (3rd row)~\citep{guo2025deepseek}. For each prompt, we iteratively append the token \textit{``Wait''} to the reasoning trace until the cumulative count of thinking tokens reaches exactly \(t_{\text{exact}}\). We consistently observe a decline in accuracy with an increase in the thinking token budget.}
\label{fig:exact_thinking}
\end{figure}

\subsection{Experimental Setup}
\paragraph{Datasets.} Following prior literature~\citep{arora2025traininglanguagemodelsreason, aggarwal2025l1}, we base our analysis on three widely-adopted mathematical reasoning datasets: (a) GSM-8K~\citep{cobbe2021gsm8k}, a dataset consisting of grade-school math problems that typically require multi-step reasoning through basic arithmetic operations to arrive at the correct answer. For this study, we evaluate on the 1,320 samples from the GSM-8K test set. (b) MATH-500~\citep{hendrycks2021measuring} is a benchmark
of challenging math problems of varying difficulty. For our study, we selected the same 500 samples selected by OpenAI in \citep{lightman2023let}, and (c) AIME 2024~\cite{AIME}, a set of $30$ competition-level problems from the American Invitational Mathematics Examination 2024, designed to assess advanced problem-solving skills across topics like arithmetic, algebra, geometry, number theory, probability, and other secondary school mathematics.

\textbf{Reasoning language models.} To ensure reproducibility, we leverage open-source reasoning models for our analysis.  Specifically, we employ three models of varying sizes and architectures, each distilled from the recently released DeepSeek-R1 model~\citep{guo2025deepseek} by \citet{guo2025deepseek}: DeepSeek-R1-Distill-Qwen-1.5B, DeepSeek-R1-Distill-Qwen-7B, and the Llama-based variant DeepSeek-R1-Distill-Llama-8B.

\textbf{Evaluation Criteria.} To evaluate reasoning performance, we report the accuracy of each model on the test set of each dataset. Specifically, for each prompt $x \in \mathcal{D}^{\text{test}}$, the model first generates a thinking trace $z$, followed by the final answer $y$. The accuracy is then defined as: $\mathbb{E}_{x\sim\mathcal{D}^{\text{test}}, z\sim\pi_\theta(\cdot\mid x),y\sim\pi_\theta(\cdot \mid x,z)}\left[\mathbb{I}\{y=y^*\}\right]$, where $y^*$ is the correct answer.

\subsection{Key Observations: Failure of Overthinking}
\label{subsec:observations}
In Figure~\ref{fig:wait_think_more}, we illustrate the trade-off between average thinking tokens and average accuracy under the \emph{Wait \& Think More} TTBC. 
We observe an initial increase (similar to~\citep{muennighoff2025s1, aggarwal2025l1}) in accuracy as the average thinking budget increases. For example, in Figure~\ref{fig:wait_think_more}(a), accuracy increases from $82.2\%$ to $87.3\%$ as the average number of thinking tokens increases from $385$ to $1100$. However, this trend does not continue indefinitely. Beyond a critical point, further increasing the thinking budget results in a steady decline in accuracy. Specifically, pushing the average thinking token count from $1100$ to $15980$ reduces accuracy from $87.3\%$ to $70.3\%$.

This observed non-monotonic trend challenges the prevailing assumption that \textit{``more thinking is always better."} Instead, it reveals a more nuanced insight: test-time reasoning exhibits a \textit{critical spot}, a point beyond which additional thinking transitions from helpful to not being helpful anymore- a phenomenon we call `overthinking'. Prior work has overlooked this degradation phase, presenting an incomplete view of the true test-time scaling landscape. To further understand and explain this phenomenon, we extend our analysis in Figure~\ref{fig:exact_thinking} using the Exact Thinking tokens setup (TTBC 2). These findings consistently reaffirm the same trend, underscoring the importance of reconsidering current test-time reasoning strategies and moving beyond the simplistic belief that more computation inherently leads to better reasoning outcomes.

In Appendix~\ref{app:min_think}, we extend this analysis to the Minimum Thinking tokens TTBC, and observe a similar non-monotonic relationship between accuracy and the number of thinking tokens.
\section{Mirage of Test-Time Scaling:  An Alternative Explanation}
\label{sec:mirage}
In this section, we propose an alternative explanation for the observed non-monotonic behavior in test-time scaling of reasoning models using a simple probabilistic framework. Although state-of-the-art reasoning models initially show accuracy improvements with increased test-time thinking, they consistently exhibit performance deterioration after exceeding a specific point. This observation prompts a fundamental question: Is the performance gain from additional thinking genuinely indicative of enhanced reasoning capability, or merely a mirage?
We illustrate the mirage effect of test-time scaling via a simple probabilistic illustration below.  

\textbf{Setup.} To illustrate this mirage effect, we consider a simplified scenario. Assume the reward is fixed and unimodal, and the model samples answers from the policy distribution $y \sim \pi(\cdot|x)$. Sepcifically, we assume $\pi(y|x) = \mathcal{N}(\mu_{\pi} , \sigma_{\pi}^2)$ (a univariate Gaussian distribution) and reward function follows a bell-curve as $r(x, y) = \mathcal{N}(\mu_r, \sigma_r^2)$. Hence, we can write
\begin{align}
    \mathbb{E}_{y \sim \pi(\cdot|x)} [r(x,y) ]= \int r(x,y) \cdot \pi(y|x)\cdot dy. 
\end{align}
From the assumptions on the policy $\pi$ and reward $r$, we can obtain 
\begin{align}
 \mathbb{E}_{y \sim \pi(\cdot|x)} [r(x,y) ] =\ &\underbrace{\frac{1}{\sqrt{2\pi(\sigma_r^2 + \sigma_\pi^2)}}}_{\text{Density Spread}} \cdot \underbrace{\exp\left( -\frac{(\mu_r - \mu_\pi)^2}{2(\sigma_r^2 + \sigma_\pi^2)} \right)}_{\text{Mean Alignment Penalty}}.\label{closed_form}
\end{align}
\begin{figure}[H]
    \centering
%
    \begin{subfigure}[b]{0.40\textwidth}
        \centering
        \includegraphics[width=\textwidth]{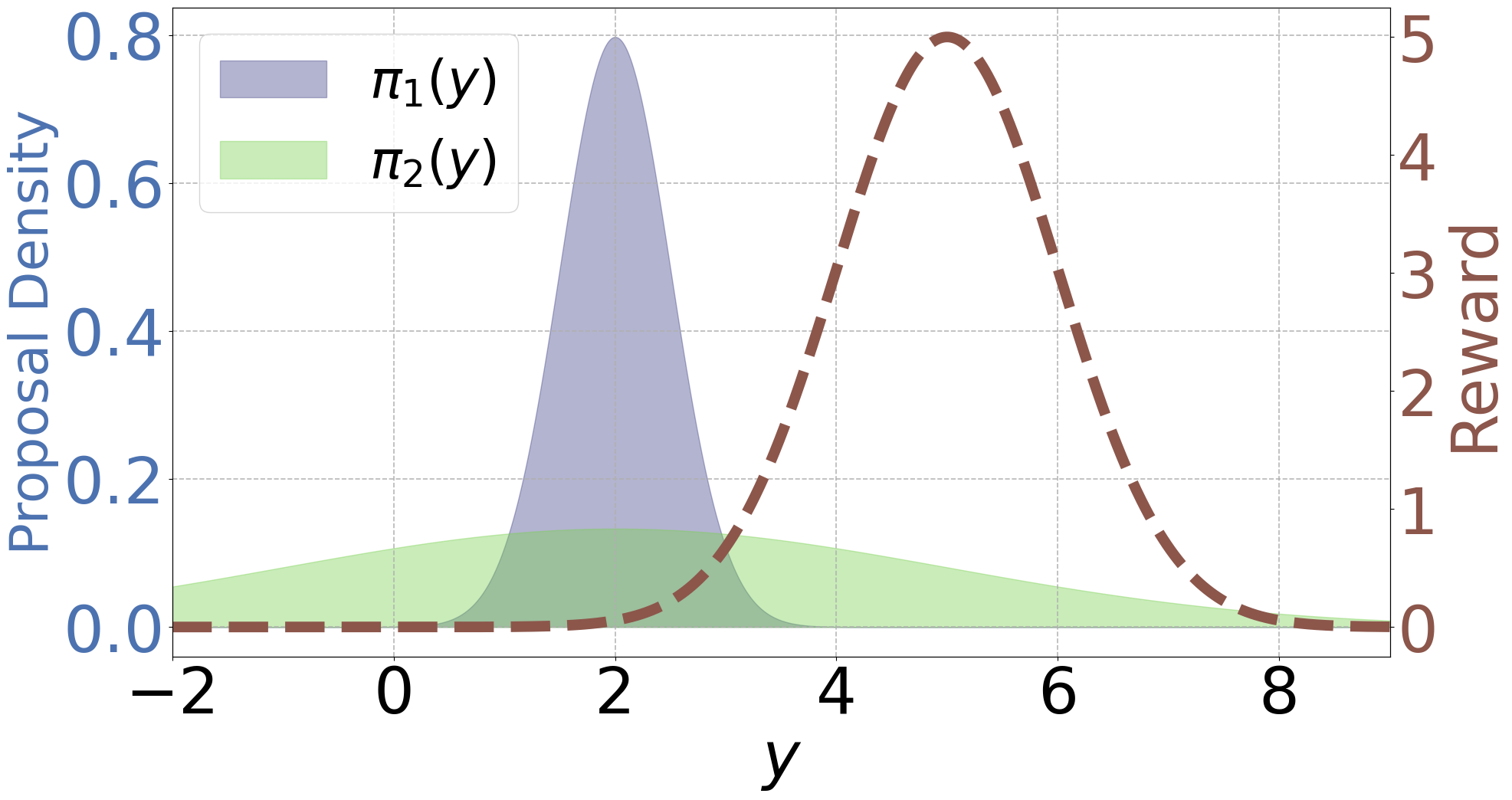}
        \caption{}
    \end{subfigure}
    \begin{subfigure}[b]{0.40\textwidth}
        \centering
        \includegraphics[width=\textwidth]{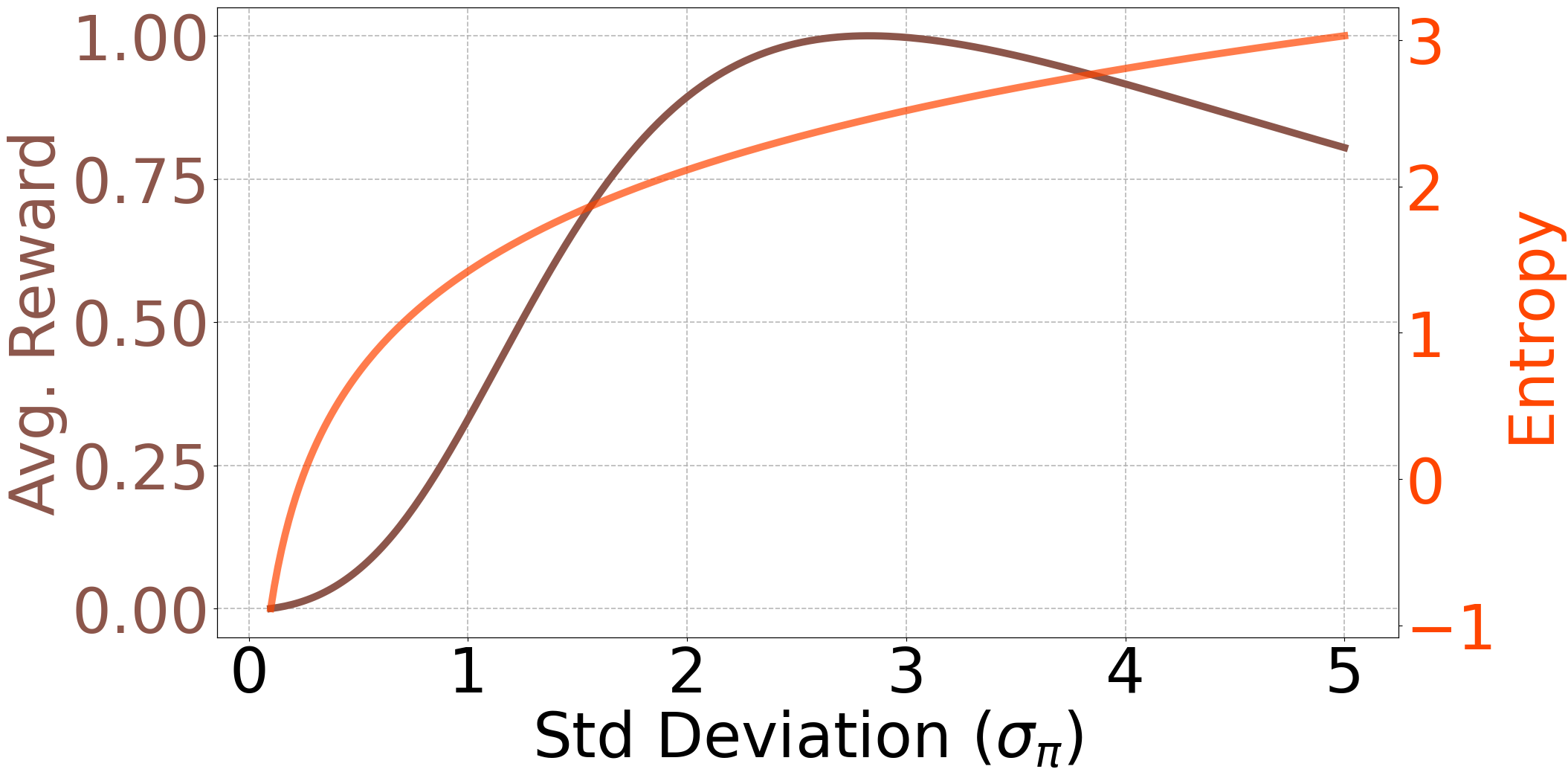}
        \caption{}
    \end{subfigure}

    \vskip\baselineskip
    \begin{subfigure}[b]{0.40\textwidth}
        \centering
        \includegraphics[width=\textwidth]{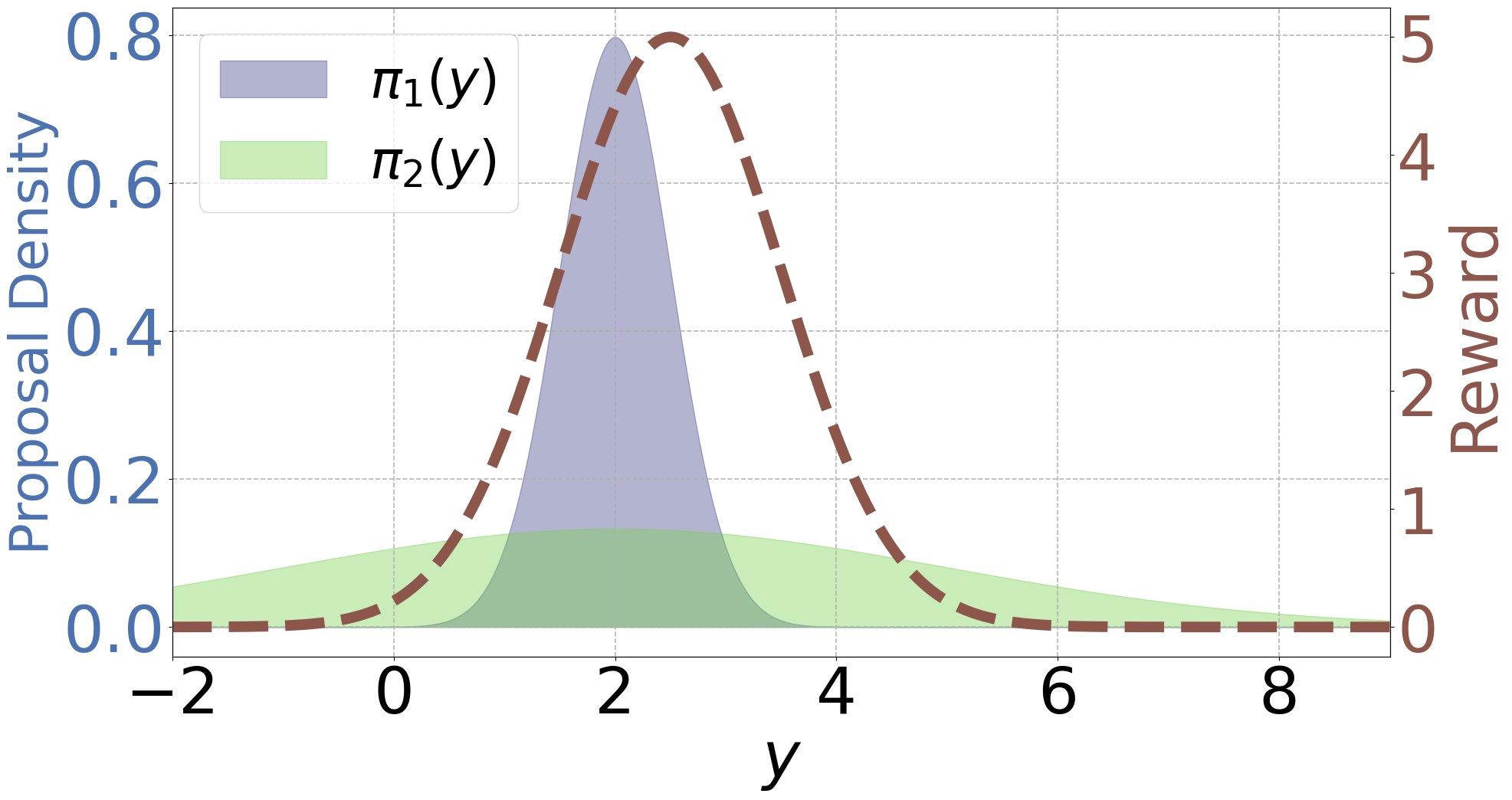}
        \caption{}
    \end{subfigure}
    \begin{subfigure}[b]{0.40\textwidth}
        \centering
        \includegraphics[width=\textwidth]{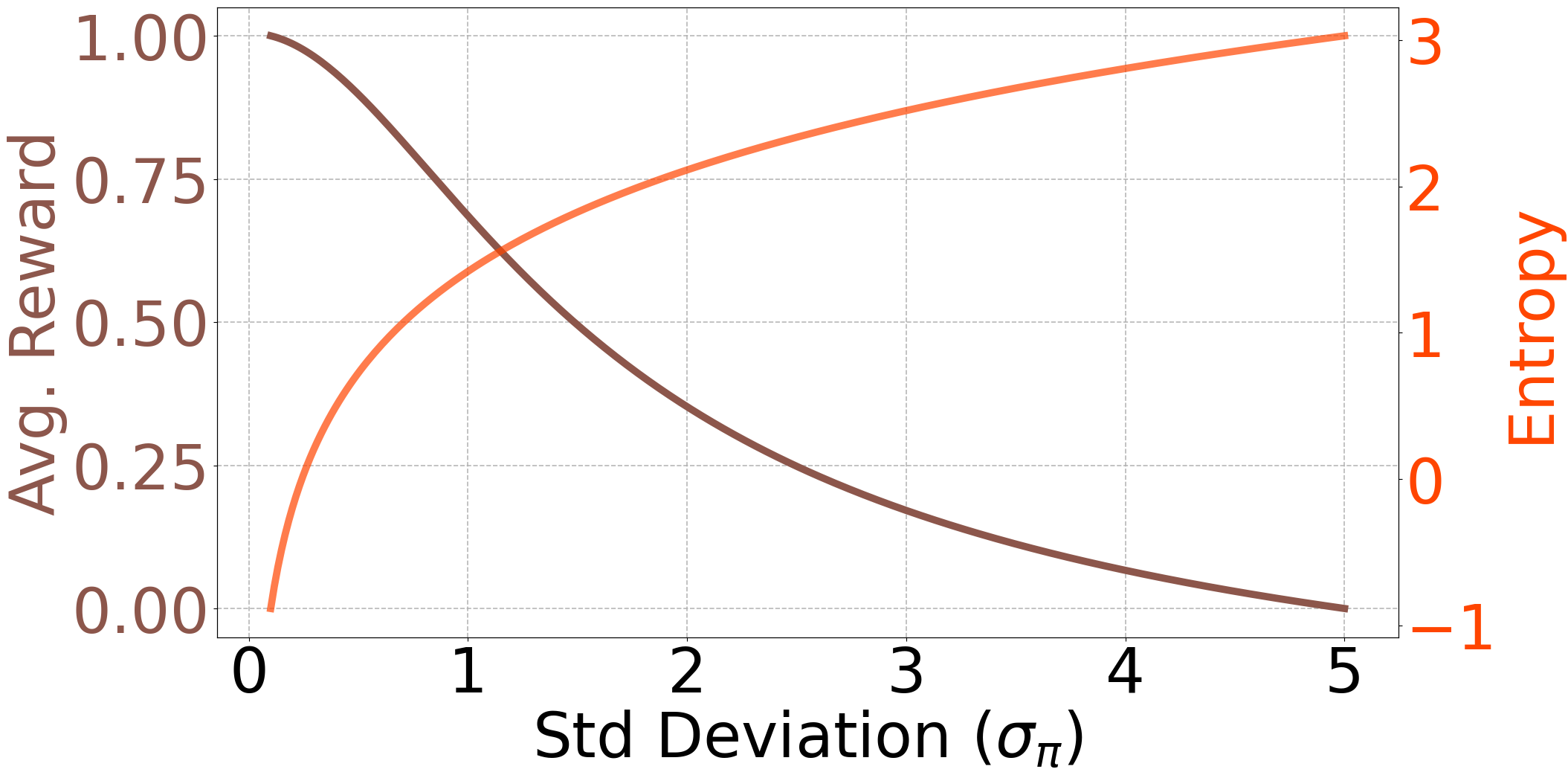}
        \caption{}
    \end{subfigure}
    \caption{\textbf{A simple illustration of the mirage in test-time scaling.}
    \textbf{Scenario 1:} (top row) When the mean of the proposal policy and reward function is far away (cf. (a)), as the variance of the proposal increases, overlap with the reward improves initially but degrades past a sweet spot, shown by the peak in reward (brown line in (b)).     \textbf{Scenario 2:} When the mean of the proposal policy and reward function is well aligned (cf. (c)), increasing variance results in diminished reward with continued entropy growth (cf. (d)).} 
    \label{fig:mirage_scaling}
\end{figure} 

\textbf{Effect of increase in variance $\sigma_{\pi}^2$.} As shown in Figures~\ref{fig:mirage_scaling}, we  observe that initially increasing $\sigma_{\pi}^2$ helps: it broadens $\pi(y|x)$, increasing its overlap with the reward distribution $r(x,y)$. As a result, the expected reward increases, leading to improved performance. However, this trend does not continue indefinitely, as performance begins to degrade beyond a certain point with more increase in variance. When the variance is too small, the model remains stuck near the proposal mean, resulting in poor reward due to limited exploration. Conversely, when the variance is too large, the model samples indiscriminately across the space, again leading to poor reward. Thus there exists a \textit{critical point} in the variance of the proposal distribution.

\textbf{Why does the critical point exist?} It arises because of two competing forces: \textit{Coverage effect:} For small $\sigma_{\pi}^2$, increasing variance improves average reward by covering more of the reward peak centered at $\mu_r$.
\textit{Dilution effect:} Beyond a point, increasing variance overspreads the distribution, placing mass on regions far from $\mu_r$, leading to diminished expected reward. The trade-off is evident in \eqref{closed_form}, 
where initially, the exponential term dominates: increasing $\sigma_\pi^2$ helps reduce the exponent (denominator increases), improving expected reward. Eventually, the prefactor ${1}/{\sqrt{2\pi(\sigma_r^2 + \sigma_\pi^2)}}$ shrinks faster than the exponent gains, reducing the overall value.

\textbf{Connection with the test-time scaling in reasoning models.} Our simple illustration reveals a powerful insight: increasing the variance of the policy distribution can initially boost expected reward, not because the policy has improved, but due to greater overlap with the reward. Crucially, this improvement is a mirage, driven by randomness rather than genuine policy refinement.  We hypothesize that a similar effect underlies the observed gains in test-time scaling of reasoning models observed in Figure \ref{fig:wait_think_more}.
Specifically, increasing thinking via longer thinking traces with prompts like ``Wait'' acts as a knob to affect the variance of the model's output distribution.

In the next subsection, we draw this analogy directly: each additional reasoning/thinking step increases the entropy of the policy, leading to broader sampling and a rise in accuracy, up to a point. Beyond that, the distribution becomes too diffuse, and performance deteriorates.

\subsection{Role of Variance in test-time scaling of reasoning models} \label{variace_role}

In this section, we empirically investigate the effect of extended thinking on model uncertainty and its role in the mirage phenomenon (Section~\ref{sec:mirage}). Specifically, our goal is to analyze how increasing the length of thinking tokens influences the entropy of the policy distribution. 

\begin{figure}[H]
    \centering
   
    \includegraphics[width=0.75\linewidth]{figures/entropy_plots_new.pdf}
    \caption{\textbf{Role of variance in test-time scaling.} We analyze how extended thinking influences policy uncertainty by visualizing entropy as a function of additional thinking tokens on two benchmarks: GSM-8K~\citep{cobbe2021gsm8k} (Figures (a) and (c)) and MATH-500~\citep{hendrycks2021measuring} (Figures (b) and (d)). We observe that with additional thinking, there is an initial rise in entropy, leading to improved performance. However, beyond a critical point, extended thinking results in a steep rise in entropy, adversely affecting performance. For this analysis, we used the ``Wait \& Think more'' setup with the DeepSeek-R1-Distill-Qwen-1.5B model.}
    \label{fig:entropy}
\end{figure}

For this analysis, we evaluate the DeepSeek-R1-Distill-Qwen-1.5B model~\citep{guo2025deepseek} on two mathematical reasoning benchmarks: GSM-8K~\citep{cobbe2021gsm8k} and MATH-500~\citep{hendrycks2021measuring}. We adopt the \emph{Wait \& Think more} setup, where we suppress the generation of the end-of-thinking delimiter and append \textit{``Wait''} to encourage the model for additional thinking. To measure the entropy of the policy, after appending each \textit{``Wait''}, we sample multiple answers $y \sim \pi(\cdot|z_{1:i}, x)$, where $z_{1:i} = [z_1, \cdots, z_{i-1},  z_i]$ denotes the cumulative sequence of thinking tokens generated up to the \(i\)th-\textit{``Wait''}, $z_1$ denotes the initial thinking trace generated with standard thinking, and $x$ is the given prompt. We quantify policy uncertainty using the entropy of the distribution over final answers: $\mathbb{E}_{y \sim \pi(\cdot|z_{1:i}, x)}[-\log \pi(y\mid z_{1:i}, x)]$. 

We report the results for this analysis averaged over all prompts in the test dataset in Figure~\ref{fig:entropy}, where baseline entropy for standard thinking (without any additional \textit{``Wait''}) is marked by a $\star$ (\textbf{\textcolor{black}{black star}}). We observe: (1) For both benchmarks, the entropy of the policy distribution increases as the model is prompted to think longer. For example, in Figure~\ref{fig:entropy} (c) for GSM-8K, extending the average thinking tokens from $385$ (standard thinking) to $6136$ resulted in a $12$x increase in entropy, from $0.23$ to $2.79$. (2) As discussed in Section~\ref{sec:mirage}, this initial rise in entropy leads to an initial improvement in average accuracy. For instance, for MATH-500 in Figure~\ref{fig:entropy} (d), the initial increase in entropy from $0.02$ to $0.12$ (with thinking extended from $2657$ to $4983$ tokens) corresponds to a $1.5\%$ accuracy gain, as shown in Figure~\ref{fig:entropy} (b).  However, with further thinking (from $4983$ to $16109$ tokens), there is a steep rise in the entropy of the policy distribution from $0.12$ to $3.0$, indicating an increase in the policy's uncertainty. This rise in entropy also correlates with a decline in average accuracy from $83.2\%$ with standard thinking to $78.3\%$ (see Figure~\ref{fig:entropy} (b)). We extend this analysis to other models in Appendix~\ref{app:extended_variance_analysis}, where we observe a similar trend of increasing entropy with additional thinking tokens.

\vspace{0.2cm}
\noindent\textbf{Key takeaway and budget control problem:} Our investigation reveals that the initial accuracy gains from test-time scaling up to some average thinking tokens are largely a mirage, arising from randomness rather than the model's refinement. This observation further implies that the \emph{``Wait \& Think More''} approach is not an optimal way to utilize the test-time thinking budget, which leads to a fundamental question: \emph{Given a fixed thinking budget, how to optimally allocate resources for test-time scaling in reasoning models?} In the next section, we provide a solution to effective budget control. 

\begin{figure*}[!t]
\centering
\begin{subfigure}{.45\textwidth}
  \centering
  \includegraphics[width=\linewidth]{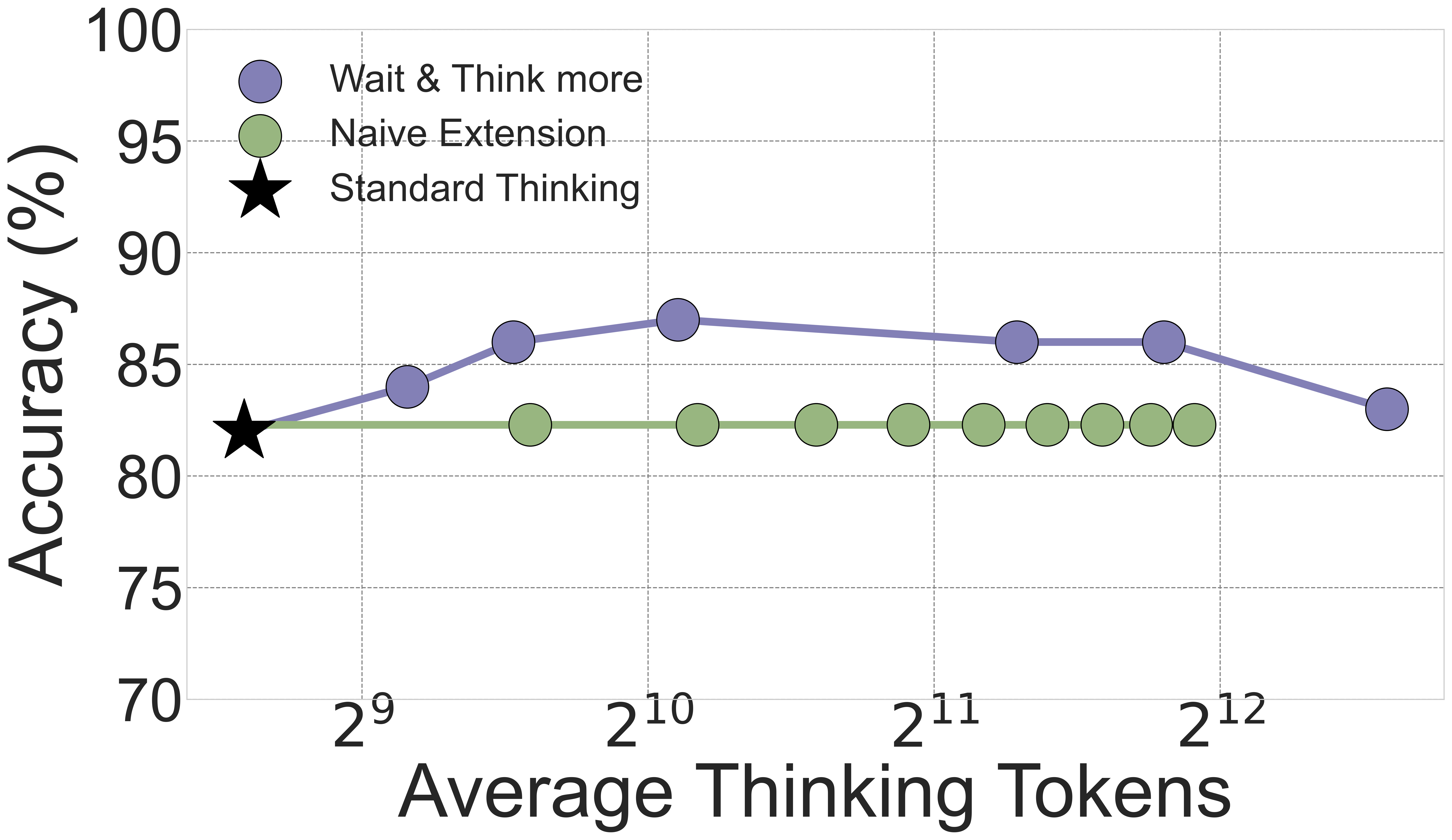}
  \caption{}
\end{subfigure} \hfill
\begin{subfigure}{.45\textwidth}
  \centering
  \includegraphics[width=\linewidth]{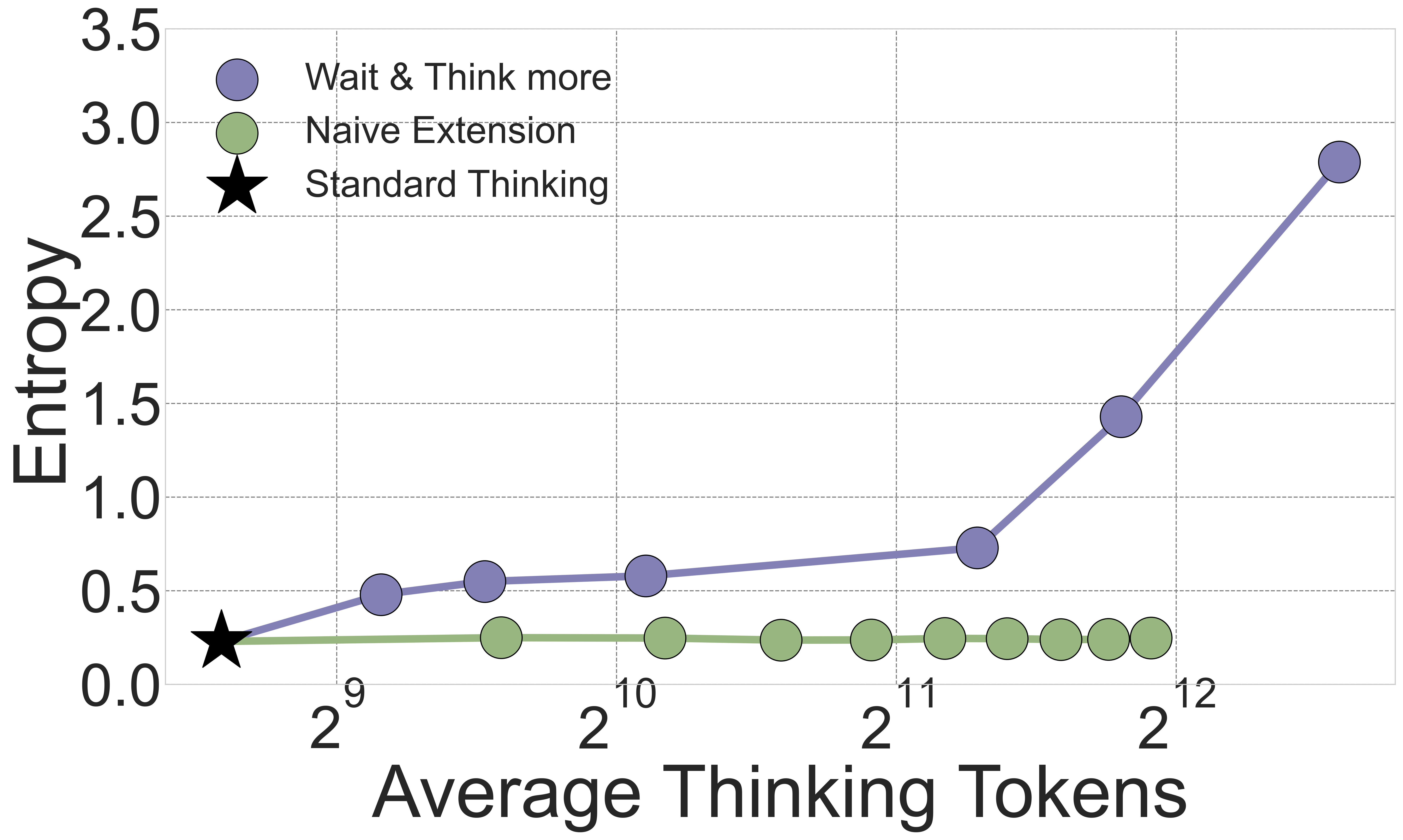}
   \caption{}
\end{subfigure}
\caption{\textbf{Understanding the impact of context length on generation.} To investigate whether increased variance originates from longer input sequences or from overthinking, we performed an experiment on GSM-8K using DeepSeek-R1-Distill-Qwen-1.5B. Rather than sampling diverse reasoning traces, we repeatedly appended the same reasoning step, $z_1 \sim \pi(\cdot \mid x)$, to artificially extend the input before generating the final answer $y \sim \pi(\cdot \mid x, z_1, z_1, \dots, z_1)$. The entropy of the resulting response distribution remained nearly unchanged, demonstrating that simply increasing context length does not induce greater uncertainty.}
\label{fig:naive_thinking}
\vspace{-0.25cm}
\end{figure*}

To gain deeper insight into the source of increased variance during overthinking, we conducted an additional experiment on GSM-8K using DeepSeek-R1-Distill-Qwen-1.5B. Specifically, instead of extending the chain-of-thought through repeated sampling of diverse reasoning traces, we artificially lengthened the reasoning sequence by repeatedly appending the same reasoning step, $z_1 \sim \pi(\cdot \mid x)$, multiple times before generating the final answer $y \sim \pi(\cdot \mid x, z_1, z_1, \dots, z_1)$. We then measured the entropy of the resulting response distribution, as reported in Figure~\ref{fig:naive_thinking} (b). The results show that entropy does not increase simply by lengthening the input with repeated tokens. This finding suggests that it is not just the context length but the overthinking that affects the performance.

\section{Effective Test-Time Budget Utilization via Parallel Thinking} \label{sec:effective_test}

\begin{figure}[!ht]
    \centering
    \includegraphics[width=0.9\linewidth]{figures/bon.pdf}
    \caption{\textbf{Comparing test-time scaling strategies.} We compare test-time scaling using \textit{``Wait \& Think more''} and parallel thinking across three benchmark datasets (GSM-8K~\citep{cobbe2021gsm8k}, MATH-500~\citep{hendrycks2021measuring}, and AIME 2024~\citep{AIME}) and three state-of-the-art reasoning models: DeepSeek-R1-Distill-Qwen-1.5B (first row), DeepSeek-R1-Distill-Qwen-7B (second row), and DeepSeek-R1-Distill-Llama-8B (third row). Each plot shows average accuracy (y-axis) against average thinking tokens (x-axis) as the thinking budget is progressively increased, with averages taken across all test set prompts. Baseline performance without budget control (standard thinking) is marked by a \textbf{\textcolor{black}{black star}}. Additional thinking using \emph{``Wait \& Think More''} is shown as \textbf{\textcolor{salmon}{salmon circles}}, while parallel scaling is represented by \textbf{\textcolor{darkgreen}{green circles}}. Unlike sequential scaling, which can degrade performance with overthinking, parallel scaling consistently maintains or improves accuracy as the budget increases.}
    \label{fig:parallel_thinking}
\end{figure}

In this section, we propose a simple approach for efficiently utilizing the given thinking token budget (say $B$ tokens) by generating multiple parallel streams of thoughts. To be specific, instead of spending the entire budget $B$ on a single extended chain of thought with $B$ tokens, we use the model $\pi(\cdot|x)$ to generate $N$ parallel thinking traces $z^{(i)} \sim \pi(\cdot \mid x), \quad i = 1, \dots, N$ such that $\sum_{i=1}^N |z^{(i)}| \leq B$, where $B$ is the thinking budget. Next, for each reasoning trace $z^{(i)}$, we sample a single output
\begin{wrapfigure}{r}{0.45\textwidth}
\vspace{-4mm}
  \begin{center}
    \includegraphics[width=0.45\textwidth]{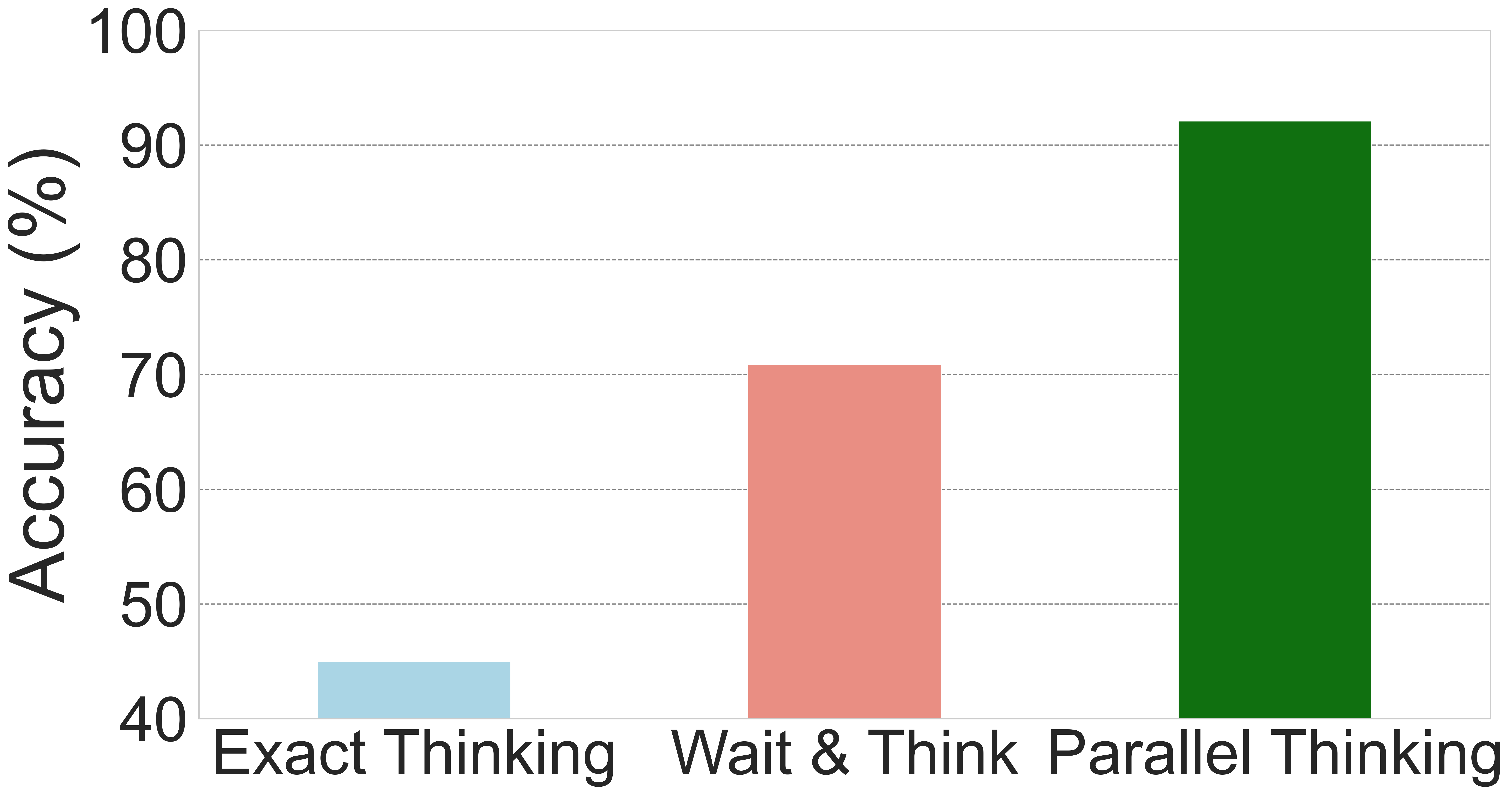}
  \end{center}
  \vspace{-4mm}
  \caption{\textbf{Effective utilization of test-time budget.} Given a fixed inference budget of $16K$ tokens, parallel thinking results in around $22\%$ and $47\%$ higher accuracy compared to \textit{Wait \& Think More} and \textit{Exact Thinking TTBC}, respectively.}
  \label{fig:budget_plot}
  \vspace{-4mm}
\end{wrapfigure} $y^{(i)} \sim \pi(\cdot \mid z^{(i)}, x)$. From the resulting candidate set $\mathcal{Y}=\{y^{(1)}, y^{(2)}, \dots, y^{(N)}\}$, we select the final output as: 
$        y^{\text{best}} = \argmax_{y \in \mathcal{Y}} R_{\text{self}} (x,y)
$, 
where  we define $R_{\text{self}} (x, y):= \sum_{i=1}^N \mathbb{I}[y^{(i)} = y]$ and $\mathbb{I}$ is the indicator function.   The rationale for employing this self-reward mechanism is to favor high-likelihood responses. While directly generating the optimal (i.e., highest probability) response from an autoregressive language model is known to be NP-hard in the worst-case scenario \cite{huang2024self}, evaluating the model's likelihood for a given response $y^{(i)}$
(conditioned on $x$ and $z^{(i)}$) is straightforward. Hence, we leverage this tractable method for identifying and selecting high-reward responses from the set of generated candidates.

\vspace{0.2cm}
\noindent\textbf{Empirical evaluations.} In Figure~\ref{fig:parallel_thinking}, we compare sequential thinking using ``Wait \& Think more'' and parallel thinking across three benchmark datasets: GSM-8K~\citep{cobbe2021gsm8k}, MATH-500~\citep{hendrycks2021measuring}, and AIME 2024~\citep{AIME}, and three state-of-the-art reasoning models. We note that, unlike sequential scaling, where extended thinking can degrade model performance, parallel thinking consistently maintains or enhances accuracy with increasing thinking budget. For example, in Figure~\ref{fig:parallel_thinking}(a), when evaluating the DeepSeek-R1-Distill-Qwen-1.5B model on GSM-8K, increasing the thinking budget from \(385\) tokens (standard thinking) to \(16000\) tokens using the \textit{``Wait \& Think More''} approach results in a \(11.8\%\) performance drop. In contrast, utilizing the same thinking token budget through parallel thinking boosts accuracy by \(10.1\%\) compared to standard thinking.
\section{Related Works}
\label{sec:related_works}
\vspace{0.2cm}
\noindent\textbf{Large reasoning models.} OpenAI's o1~\citep{openo1} series introduced the first large-scale language model to integrate RL-driven training with the concept of test-time thinking, unlocking the reasoning potential of LLMs, and establishing the domain of \emph{large reasoning models}~\citep{xu2025towards}. Following the release of o1, numerous efforts have sought to replicate its reasoning capabilities, including Open-R~\citep{wang2024openr}, Llama-berry~\citep{zhang2024llama}, Journey-R~\citep{qin2024o1}, and Drt-o1~\citep{wang2024drt}. These approaches typically leverage Monte-Carlo Tree Search (MCTS)~\citep{gao2024interpretable, qin2024o1, zhao2024marco, zhang2024o1, zhang2024llama} or distillation from reasoning traces generated by o1-like models~\citep{huang2024o1}. One of the first open-source models to match o1-level reasoning performance was DeepSeek-R1~\citep{guo2025deepseek}, which employed a sparse-reward based RL training framework~\citep{shao2024deepseekmath} on the DeepSeek-V3-Base model~\citep{deepseekai2024deepseekv3technicalreport} to enhance its reasoning capabilities. Following DeepSeek-R1,  recently several open-source efforts have continued employing RL training~\citep{shao2024deepseekmath} to improve reasoning performance~\citep{qwq32b, arora2025traininglanguagemodelsreason, aggarwal2025l1, liu2025visual, shen2025vlm, yu2025dapo, huang2025vision, jin2025search, team2025kimi, xai2025grok35, deepmind2024gemini25}. In this study, we specifically focus on analyzing the test-time scaling behavior in reasoning models.

\vspace{0.3cm}
\noindent\textbf{Test-time scaling in reasoning language models.} Recent work by \citet{muennighoff2025s1} introduced the concept of budget forcing to replicate the test-time scaling behavior observed in o1 models~\citep{openo1}. Another recent approach, L1~\citep{aggarwal2025l1}, proposed length-controlled policy optimization, providing precise control over the length of the reasoning trace during generation. \citet{yang2025thinkingoptimalscalingtesttimecompute} introduced a thinking-optimal scaling strategy, training models to adapt dynamically to different levels of reasoning effort depending on the test-time compute budget. Recently, a lot of studies have also focused on fine-tuning models to think efficiently according to task complexity~\citep{arora2025traininglanguagemodelsreason, fang2025thinkless, zhang2025continue, jiang2025think, liang2025thinkswitcher, zhang2025adaptthink, huang2025adactrl}.

\section{Conclusions and Limitations}

In this paper, we present a systematic study of test-time scaling in reasoning models, analyzing how prompting models to \textit{“Wait \& think”} affects performance. Our key finding is that while extended thinking initially improves accuracy, continued reasoning beyond a threshold degrades performance—a phenomenon we term \textit{overthinking}. We show that these early gains are often illusory, stemming from increased response variance rather than genuine reasoning improvements. To address this, we propose a simple and effective alternative: \textit{parallel thinking}, which distributes the inference budget across multiple reasoning paths and selects the final answer via majority vote, yielding consistent gains.

\noindent\textbf{Limitations.} Our study is limited to mid-sized models due to resource constraints; evaluating the behavior of larger models (e.g., 32B and 70B) remains an exciting avenue for future work. Moreover, while our insights are supported by empirical evidence and a simple probabilistic model, developing a formal theoretical understanding of how variance and entropy affect reasoning performance is an important direction for deeper investigation.

\bibliography{main}

\begin{thebibliography}{67}
\providecommand{\natexlab}[1]{#1}
\providecommand{\url}[1]{\texttt{#1}}
\expandafter\ifx\csname urlstyle\endcsname\relax
  \providecommand{\doi}[1]{doi: #1}\else
  \providecommand{\doi}{doi: \begingroup \urlstyle{rm}\Url}\fi

\bibitem[Aggarwal \& Welleck(2025)Aggarwal and Welleck]{aggarwal2025l1}
Pranjal Aggarwal and Sean Welleck.
\newblock L1: Controlling how long a reasoning model thinks with reinforcement learning.
\newblock \emph{arXiv preprint arXiv:2503.04697}, 2025.

\bibitem[Amini et~al.(2024)Amini, Vieira, Ash, and Cotterell]{amini2024variational}
Afra Amini, Tim Vieira, Elliott Ash, and Ryan Cotterell.
\newblock Variational best-of-n alignment.
\newblock \emph{arXiv preprint arXiv:2407.06057}, 2024.

\bibitem[Arora \& Zanette(2025)Arora and Zanette]{arora2025traininglanguagemodelsreason}
Daman Arora and Andrea Zanette.
\newblock Training language models to reason efficiently.
\newblock 2025.
\newblock URL \url{https://arxiv.org/abs/2502.04463}.

\bibitem[Azerbayev et~al.(2023)Azerbayev, Schoelkopf, Paster, Santos, McAleer, Jiang, Deng, Biderman, and Welleck]{azerbayev2023llemma}
Zhangir Azerbayev, Hailey Schoelkopf, Keiran Paster, Marco~Dos Santos, Stephen McAleer, Albert~Q Jiang, Jia Deng, Stella Biderman, and Sean Welleck.
\newblock Llemma: An open language model for mathematics.
\newblock \emph{arXiv preprint arXiv:2310.10631}, 2023.

\bibitem[Beirami et~al.(2024)Beirami, Agarwal, Berant, D'Amour, Eisenstein, Nagpal, and Suresh]{beirami2024theoretical}
Ahmad Beirami, Alekh Agarwal, Jonathan Berant, Alexander D'Amour, Jacob Eisenstein, Chirag Nagpal, and Ananda~Theertha Suresh.
\newblock Theoretical guarantees on the best-of-n alignment policy.
\newblock \emph{arXiv preprint arXiv:2401.01879}, 2024.

\bibitem[Cobbe et~al.(2021)Cobbe, Kosaraju, Bavarian, Chen, Jun, Kaiser, Plappert, Tworek, Hilton, Nakano, Hesse, and Schulman]{cobbe2021gsm8k}
Karl Cobbe, Vineet Kosaraju, Mohammad Bavarian, Mark Chen, Heewoo Jun, Lukasz Kaiser, Matthias Plappert, Jerry Tworek, Jacob Hilton, Reiichiro Nakano, Christopher Hesse, and John Schulman.
\newblock Training verifiers to solve math word problems.
\newblock \emph{arXiv preprint arXiv:2110.14168}, 2021.

\bibitem[DeepSeek-AI(2024)]{deepseekai2024deepseekv3technicalreport}
DeepSeek-AI.
\newblock Deepseek-v3 technical report, 2024.
\newblock URL \url{https://arxiv.org/abs/2412.19437}.

\bibitem[Fang et~al.(2025)Fang, Ma, and Wang]{fang2025thinkless}
Gongfan Fang, Xinyin Ma, and Xinchao Wang.
\newblock Thinkless: Llm learns when to think.
\newblock \emph{arXiv preprint arXiv:2505.13379}, 2025.

\bibitem[Gao et~al.(2024)Gao, Niu, He, Xu, Liu, Liu, Hu, and Wen]{gao2024interpretable}
Zitian Gao, Boye Niu, Xuzheng He, Haotian Xu, Hongzhang Liu, Aiwei Liu, Xuming Hu, and Lijie Wen.
\newblock Interpretable contrastive monte carlo tree search reasoning.
\newblock \emph{arXiv preprint arXiv:2410.01707}, 2024.

\bibitem[{Google DeepMind}(2024)]{deepmind2024gemini25}
{Google DeepMind}.
\newblock {Gemini 2.5 Pro: The latest Gemini multimodal model}.
\newblock \url{https://deepmind.google/technologies/gemini/}, may 2024.

\bibitem[Gui et~al.(2024)Gui, Garbacea, and Veitch]{bon3}
Lin Gui, Cristina Garbacea, and Victor Veitch.
\newblock Bonbon alignment for large language models and the sweetness of best-of-n sampling.
\newblock In \emph{The Thirty-eighth Annual Conference on Neural Information Processing Systems}, 2024.

\bibitem[Guo et~al.(2025)Guo, Yang, Zhang, Song, Zhang, Xu, Zhu, Ma, Wang, Bi, et~al.]{guo2025deepseek}
Daya Guo, Dejian Yang, Haowei Zhang, Junxiao Song, Ruoyu Zhang, Runxin Xu, Qihao Zhu, Shirong Ma, Peiyi Wang, Xiao Bi, et~al.
\newblock Deepseek-r1: Incentivizing reasoning capability in llms via reinforcement learning.
\newblock \emph{arXiv preprint arXiv:2501.12948}, 2025.

\bibitem[Hendrycks et~al.(2021{\natexlab{a}})Hendrycks, Basart, Kadavath, Mazeika, Arora, Guo, Burns, Puranik, He, Song, et~al.]{hendrycks2021measuringcode}
Dan Hendrycks, Steven Basart, Saurav Kadavath, Mantas Mazeika, Akul Arora, Ethan Guo, Collin Burns, Samir Puranik, Horace He, Dawn Song, et~al.
\newblock Measuring coding challenge competence with apps.
\newblock \emph{arXiv preprint arXiv:2105.09938}, 2021{\natexlab{a}}.

\bibitem[Hendrycks et~al.(2021{\natexlab{b}})Hendrycks, Burns, Kadavath, Arora, Basart, Tang, Song, and Steinhardt]{hendrycks2021measuring}
Dan Hendrycks, Collin Burns, Saurav Kadavath, Akul Arora, Steven Basart, Eric Tang, Dawn Song, and Jacob Steinhardt.
\newblock Measuring mathematical problem solving with the math dataset.
\newblock \emph{arXiv preprint arXiv:2103.03874}, 2021{\natexlab{b}}.

\bibitem[Hou et~al.(2025)Hou, Zhang, Ji, Liu, Qian, Andreas, and Chang]{hou2025thinkprune}
Bairu Hou, Yang Zhang, Jiabao Ji, Yujian Liu, Kaizhi Qian, Jacob Andreas, and Shiyu Chang.
\newblock Thinkprune: Pruning long chain-of-thought of llms via reinforcement learning.
\newblock \emph{arXiv preprint arXiv:2504.01296}, 2025.

\bibitem[Huang et~al.(2024{\natexlab{a}})Huang, Block, Foster, Rohatgi, Zhang, Simchowitz, Ash, and Krishnamurthy]{huang2024self}
Audrey Huang, Adam Block, Dylan~J Foster, Dhruv Rohatgi, Cyril Zhang, Max Simchowitz, Jordan~T Ash, and Akshay Krishnamurthy.
\newblock Self-improvement in language models: The sharpening mechanism.
\newblock \emph{arXiv preprint arXiv:2412.01951}, 2024{\natexlab{a}}.

\bibitem[Huang et~al.(2025{\natexlab{a}})Huang, Wang, Zhong, Su, Feng, Cao, and Fung]{huang2025adactrl}
Shijue Huang, Hongru Wang, Wanjun Zhong, Zhaochen Su, Jiazhan Feng, Bowen Cao, and Yi~R Fung.
\newblock Adactrl: Towards adaptive and controllable reasoning via difficulty-aware budgeting.
\newblock \emph{arXiv preprint arXiv:2505.18822}, 2025{\natexlab{a}}.

\bibitem[Huang et~al.(2025{\natexlab{b}})Huang, Jia, Zhai, Cao, Ye, Zhao, Xu, Hu, and Lin]{huang2025vision}
Wenxuan Huang, Bohan Jia, Zijie Zhai, Shaosheng Cao, Zheyu Ye, Fei Zhao, Zhe Xu, Yao Hu, and Shaohui Lin.
\newblock Vision-r1: Incentivizing reasoning capability in multimodal large language models.
\newblock \emph{arXiv preprint arXiv:2503.06749}, 2025{\natexlab{b}}.

\bibitem[Huang et~al.(2024{\natexlab{b}})Huang, Zou, Li, Liu, Zheng, Chern, Xia, Qin, Yuan, and Liu]{huang2024o1}
Zhen Huang, Haoyang Zou, Xuefeng Li, Yixiu Liu, Yuxiang Zheng, Ethan Chern, Shijie Xia, Yiwei Qin, Weizhe Yuan, and Pengfei Liu.
\newblock O1 replication journey--part 2: Surpassing o1-preview through simple distillation, big progress or bitter lesson?
\newblock \emph{arXiv preprint arXiv:2411.16489}, 2024{\natexlab{b}}.

\bibitem[invitational~mathematics examination(2024)]{AIME}
American invitational~mathematics examination.
\newblock Amc.
\newblock 2024.
\newblock URL \url{https://artofproblemsolving. com/wiki/index.php/American Invitational Mathematics Examination}.

\bibitem[Jiang et~al.(2025)Jiang, Wu, Huang, Dong, Chi, Dong, Zhang, Lv, Cui, and Wei]{jiang2025think}
Lingjie Jiang, Xun Wu, Shaohan Huang, Qingxiu Dong, Zewen Chi, Li~Dong, Xingxing Zhang, Tengchao Lv, Lei Cui, and Furu Wei.
\newblock Think only when you need with large hybrid-reasoning models.
\newblock \emph{arXiv preprint arXiv:2505.14631}, 2025.

\bibitem[Jin et~al.(2025)Jin, Zeng, Yue, Yoon, Arik, Wang, Zamani, and Han]{jin2025search}
Bowen Jin, Hansi Zeng, Zhenrui Yue, Jinsung Yoon, Sercan Arik, Dong Wang, Hamed Zamani, and Jiawei Han.
\newblock Search-r1: Training llms to reason and leverage search engines with reinforcement learning.
\newblock \emph{arXiv preprint arXiv:2503.09516}, 2025.

\bibitem[Jinnai et~al.(2024)Jinnai, Morimura, Ariu, and Abe]{bon4}
Yuu Jinnai, Tetsuro Morimura, Kaito Ariu, and Kenshi Abe.
\newblock Regularized best-of-n sampling to mitigate reward hacking for language model alignment, 2024.
\newblock URL \url{https://arxiv.org/abs/2404.01054}.

\bibitem[Kumar et~al.(2024)Kumar, Zhuang, Agarwal, Su, Co-Reyes, Singh, Baumli, Iqbal, Bishop, Roelofs, et~al.]{kumar2024training}
Aviral Kumar, Vincent Zhuang, Rishabh Agarwal, Yi~Su, John~D Co-Reyes, Avi Singh, Kate Baumli, Shariq Iqbal, Colton Bishop, Rebecca Roelofs, et~al.
\newblock Training language models to self-correct via reinforcement learning.
\newblock \emph{arXiv preprint arXiv:2409.12917}, 2024.

\bibitem[Lewkowycz et~al.(2022)Lewkowycz, Andreassen, Dohan, Dyer, Michalewski, Ramasesh, Slone, Anil, Schlag, Gutman-Solo, et~al.]{lewkowycz2022solving}
Aitor Lewkowycz, Anders Andreassen, David Dohan, Ethan Dyer, Henryk Michalewski, Vinay Ramasesh, Ambrose Slone, Cem Anil, Imanol Schlag, Theo Gutman-Solo, et~al.
\newblock Solving quantitative reasoning problems with language models.
\newblock \emph{Advances in Neural Information Processing Systems}, 35:\penalty0 3843--3857, 2022.

\bibitem[Li et~al.(2024)Li, Cui, Zhao, Kong, and Bi]{li2024gsm}
Qintong Li, Leyang Cui, Xueliang Zhao, Lingpeng Kong, and Wei Bi.
\newblock Gsm-plus: A comprehensive benchmark for evaluating the robustness of llms as mathematical problem solvers.
\newblock \emph{arXiv preprint arXiv:2402.19255}, 2024.

\bibitem[Liang et~al.(2025)Liang, Zhong, Yang, and Quan]{liang2025thinkswitcher}
Guosheng Liang, Longguang Zhong, Ziyi Yang, and Xiaojun Quan.
\newblock Thinkswitcher: When to think hard, when to think fast.
\newblock \emph{arXiv preprint arXiv:2505.14183}, 2025.

\bibitem[Lightman et~al.(2023)Lightman, Kosaraju, Burda, Edwards, Baker, Lee, Leike, Schulman, Sutskever, and Cobbe]{lightman2023let}
Hunter Lightman, Vineet Kosaraju, Yuri Burda, Harrison Edwards, Bowen Baker, Teddy Lee, Jan Leike, John Schulman, Ilya Sutskever, and Karl Cobbe.
\newblock Let's verify step by step.
\newblock In \emph{The Twelfth International Conference on Learning Representations}, 2023.

\bibitem[Liu et~al.(2025)Liu, Sun, Zang, Dong, Cao, Duan, Lin, and Wang]{liu2025visual}
Ziyu Liu, Zeyi Sun, Yuhang Zang, Xiaoyi Dong, Yuhang Cao, Haodong Duan, Dahua Lin, and Jiaqi Wang.
\newblock Visual-rft: Visual reinforcement fine-tuning.
\newblock \emph{arXiv preprint arXiv:2503.01785}, 2025.

\bibitem[Marjanovi{\'c} et~al.(2025)Marjanovi{\'c}, Patel, Adlakha, Aghajohari, BehnamGhader, Bhatia, Khandelwal, Kraft, Krojer, L{\`u}, et~al.]{marjanovic2025deepseek}
Sara~Vera Marjanovi{\'c}, Arkil Patel, Vaibhav Adlakha, Milad Aghajohari, Parishad BehnamGhader, Mehar Bhatia, Aditi Khandelwal, Austin Kraft, Benno Krojer, Xing~Han L{\`u}, et~al.
\newblock Deepseek-r1 thoughtology: Let's< think> about llm reasoning.
\newblock \emph{arXiv preprint arXiv:2504.07128}, 2025.

\bibitem[Muennighoff et~al.(2025)Muennighoff, Yang, Shi, Li, Fei-Fei, Hajishirzi, Zettlemoyer, Liang, Cand{\`e}s, and Hashimoto]{muennighoff2025s1}
Niklas Muennighoff, Zitong Yang, Weijia Shi, Xiang~Lisa Li, Li~Fei-Fei, Hannaneh Hajishirzi, Luke Zettlemoyer, Percy Liang, Emmanuel Cand{\`e}s, and Tatsunori Hashimoto.
\newblock s1: Simple test-time scaling.
\newblock \emph{arXiv preprint arXiv:2501.19393}, 2025.

\bibitem[Nakano et~al.(2021)Nakano, Hilton, Balaji, Wu, Ouyang, Kim, Hesse, Jain, Kosaraju, Saunders, et~al.]{nakano2021webgpt}
Reiichiro Nakano, Jacob Hilton, Suchir Balaji, Jeff Wu, Long Ouyang, Christina Kim, Christopher Hesse, Shantanu Jain, Vineet Kosaraju, William Saunders, et~al.
\newblock Webgpt: Browser-assisted question-answering with human feedback.
\newblock \emph{arXiv preprint arXiv:2112.09332}, 2021.

\bibitem[OpenAI et~al.(2024)OpenAI, Achiam, Adler, Agarwal, Ahmad, Akkaya, Aleman, Almeida, Altenschmidt, Altman, Anadkat, Avila, Babuschkin, Balaji, Balcom, Baltescu, Bao, Bavarian, Belgum, Bello, Berdine, Bernadett-Shapiro, Berner, Bogdonoff, Boiko, Boyd, Brakman, Brockman, Brooks, Brundage, Button, Cai, Campbell, Cann, Carey, Carlson, Carmichael, Chan, Chang, Chantzis, Chen, Chen, Chen, Chen, Chen, Chess, Cho, Chu, Chung, Cummings, Currier, Dai, Decareaux, Degry, Deutsch, Deville, Dhar, Dohan, Dowling, Dunning, Ecoffet, Eleti, Eloundou, Farhi, Fedus, Felix, Fishman, Forte, Fulford, Gao, Georges, Gibson, Goel, Gogineni, Goh, Gontijo-Lopes, Gordon, Grafstein, Gray, Greene, Gross, Gu, Guo, Hallacy, Han, Harris, He, Heaton, Heidecke, Hesse, Hickey, Hickey, Hoeschele, Houghton, Hsu, Hu, Hu, Huizinga, Jain, Jain, Jang, Jiang, Jiang, Jin, Jin, Jomoto, Jonn, Jun, Kaftan, Łukasz Kaiser, Kamali, Kanitscheider, Keskar, Khan, Kilpatrick, Kim, Kim, Kim, Kirchner, Kiros, Knight, Kokotajlo, Łukasz Kondraciuk, Kondrich,
  Konstantinidis, Kosic, Krueger, Kuo, Lampe, Lan, Lee, Leike, Leung, Levy, Li, Lim, Lin, Lin, Litwin, Lopez, Lowe, Lue, Makanju, Malfacini, Manning, Markov, Markovski, Martin, Mayer, Mayne, McGrew, McKinney, McLeavey, McMillan, McNeil, Medina, Mehta, Menick, Metz, Mishchenko, Mishkin, Monaco, Morikawa, Mossing, Mu, Murati, Murk, Mély, Nair, Nakano, Nayak, Neelakantan, Ngo, Noh, Ouyang, O'Keefe, Pachocki, Paino, Palermo, Pantuliano, Parascandolo, Parish, Parparita, Passos, Pavlov, Peng, Perelman, de~Avila Belbute~Peres, Petrov, de~Oliveira~Pinto, Michael, Pokorny, Pokrass, Pong, Powell, Power, Power, Proehl, Puri, Radford, Rae, Ramesh, Raymond, Real, Rimbach, Ross, Rotsted, Roussez, Ryder, Saltarelli, Sanders, Santurkar, Sastry, Schmidt, Schnurr, Schulman, Selsam, Sheppard, Sherbakov, Shieh, Shoker, Shyam, Sidor, Sigler, Simens, Sitkin, Slama, Sohl, Sokolowsky, Song, Staudacher, Such, Summers, Sutskever, Tang, Tezak, Thompson, Tillet, Tootoonchian, Tseng, Tuggle, Turley, Tworek, Uribe, Vallone, Vijayvergiya,
  Voss, Wainwright, Wang, Wang, Wang, Ward, Wei, Weinmann, Welihinda, Welinder, Weng, Weng, Wiethoff, Willner, Winter, Wolrich, Wong, Workman, Wu, Wu, Wu, Xiao, Xu, Yoo, Yu, Yuan, Zaremba, Zellers, Zhang, Zhang, Zhao, Zheng, Zhuang, Zhuk, and Zoph]{openai2024gpt4}
OpenAI, Josh Achiam, Steven Adler, Sandhini Agarwal, Lama Ahmad, Ilge Akkaya, Florencia~Leoni Aleman, Diogo Almeida, Janko Altenschmidt, Sam Altman, Shyamal Anadkat, Red Avila, Igor Babuschkin, Suchir Balaji, Valerie Balcom, Paul Baltescu, Haiming Bao, Mohammad Bavarian, Jeff Belgum, Irwan Bello, Jake Berdine, Gabriel Bernadett-Shapiro, Christopher Berner, Lenny Bogdonoff, Oleg Boiko, Madelaine Boyd, Anna-Luisa Brakman, Greg Brockman, Tim Brooks, Miles Brundage, Kevin Button, Trevor Cai, Rosie Campbell, Andrew Cann, Brittany Carey, Chelsea Carlson, Rory Carmichael, Brooke Chan, Che Chang, Fotis Chantzis, Derek Chen, Sully Chen, Ruby Chen, Jason Chen, Mark Chen, Ben Chess, Chester Cho, Casey Chu, Hyung~Won Chung, Dave Cummings, Jeremiah Currier, Yunxing Dai, Cory Decareaux, Thomas Degry, Noah Deutsch, Damien Deville, Arka Dhar, David Dohan, Steve Dowling, Sheila Dunning, Adrien Ecoffet, Atty Eleti, Tyna Eloundou, David Farhi, Liam Fedus, Niko Felix, Simón~Posada Fishman, Juston Forte, Isabella Fulford, Leo
  Gao, Elie Georges, Christian Gibson, Vik Goel, Tarun Gogineni, Gabriel Goh, Rapha Gontijo-Lopes, Jonathan Gordon, Morgan Grafstein, Scott Gray, Ryan Greene, Joshua Gross, Shixiang~Shane Gu, Yufei Guo, Chris Hallacy, Jesse Han, Jeff Harris, Yuchen He, Mike Heaton, Johannes Heidecke, Chris Hesse, Alan Hickey, Wade Hickey, Peter Hoeschele, Brandon Houghton, Kenny Hsu, Shengli Hu, Xin Hu, Joost Huizinga, Shantanu Jain, Shawn Jain, Joanne Jang, Angela Jiang, Roger Jiang, Haozhun Jin, Denny Jin, Shino Jomoto, Billie Jonn, Heewoo Jun, Tomer Kaftan, Łukasz Kaiser, Ali Kamali, Ingmar Kanitscheider, Nitish~Shirish Keskar, Tabarak Khan, Logan Kilpatrick, Jong~Wook Kim, Christina Kim, Yongjik Kim, Jan~Hendrik Kirchner, Jamie Kiros, Matt Knight, Daniel Kokotajlo, Łukasz Kondraciuk, Andrew Kondrich, Aris Konstantinidis, Kyle Kosic, Gretchen Krueger, Vishal Kuo, Michael Lampe, Ikai Lan, Teddy Lee, Jan Leike, Jade Leung, Daniel Levy, Chak~Ming Li, Rachel Lim, Molly Lin, Stephanie Lin, Mateusz Litwin, Theresa Lopez, Ryan
  Lowe, Patricia Lue, Anna Makanju, Kim Malfacini, Sam Manning, Todor Markov, Yaniv Markovski, Bianca Martin, Katie Mayer, Andrew Mayne, Bob McGrew, Scott~Mayer McKinney, Christine McLeavey, Paul McMillan, Jake McNeil, David Medina, Aalok Mehta, Jacob Menick, Luke Metz, Andrey Mishchenko, Pamela Mishkin, Vinnie Monaco, Evan Morikawa, Daniel Mossing, Tong Mu, Mira Murati, Oleg Murk, David Mély, Ashvin Nair, Reiichiro Nakano, Rajeev Nayak, Arvind Neelakantan, Richard Ngo, Hyeonwoo Noh, Long Ouyang, Cullen O'Keefe, Jakub Pachocki, Alex Paino, Joe Palermo, Ashley Pantuliano, Giambattista Parascandolo, Joel Parish, Emy Parparita, Alex Passos, Mikhail Pavlov, Andrew Peng, Adam Perelman, Filipe de~Avila Belbute~Peres, Michael Petrov, Henrique~Ponde de~Oliveira~Pinto, Michael, Pokorny, Michelle Pokrass, Vitchyr~H. Pong, Tolly Powell, Alethea Power, Boris Power, Elizabeth Proehl, Raul Puri, Alec Radford, Jack Rae, Aditya Ramesh, Cameron Raymond, Francis Real, Kendra Rimbach, Carl Ross, Bob Rotsted, Henri Roussez,
  Nick Ryder, Mario Saltarelli, Ted Sanders, Shibani Santurkar, Girish Sastry, Heather Schmidt, David Schnurr, John Schulman, Daniel Selsam, Kyla Sheppard, Toki Sherbakov, Jessica Shieh, Sarah Shoker, Pranav Shyam, Szymon Sidor, Eric Sigler, Maddie Simens, Jordan Sitkin, Katarina Slama, Ian Sohl, Benjamin Sokolowsky, Yang Song, Natalie Staudacher, Felipe~Petroski Such, Natalie Summers, Ilya Sutskever, Jie Tang, Nikolas Tezak, Madeleine~B. Thompson, Phil Tillet, Amin Tootoonchian, Elizabeth Tseng, Preston Tuggle, Nick Turley, Jerry Tworek, Juan Felipe~Cerón Uribe, Andrea Vallone, Arun Vijayvergiya, Chelsea Voss, Carroll Wainwright, Justin~Jay Wang, Alvin Wang, Ben Wang, Jonathan Ward, Jason Wei, CJ~Weinmann, Akila Welihinda, Peter Welinder, Jiayi Weng, Lilian Weng, Matt Wiethoff, Dave Willner, Clemens Winter, Samuel Wolrich, Hannah Wong, Lauren Workman, Sherwin Wu, Jeff Wu, Michael Wu, Kai Xiao, Tao Xu, Sarah Yoo, Kevin Yu, Qiming Yuan, Wojciech Zaremba, Rowan Zellers, Chong Zhang, Marvin Zhang, Shengjia
  Zhao, Tianhao Zheng, Juntang Zhuang, William Zhuk, and Barret Zoph.
\newblock Gpt-4 technical report, 2024.

\bibitem[OpenAIo1(2024)]{openo1}
OpenAIo1.
\newblock Learning to reason with llms.
\newblock 2024.
\newblock URL \url{https://openai.com/index/learning-to-reason-with-llms/}.

\bibitem[Qin et~al.(2024)Qin, Li, Zou, Liu, Xia, Huang, Ye, Yuan, Liu, Li, et~al.]{qin2024o1}
Yiwei Qin, Xuefeng Li, Haoyang Zou, Yixiu Liu, Shijie Xia, Zhen Huang, Yixin Ye, Weizhe Yuan, Hector Liu, Yuanzhi Li, et~al.
\newblock O1 replication journey: A strategic progress report--part 1.
\newblock \emph{arXiv preprint arXiv:2410.18982}, 2024.

\bibitem[Ramesh et~al.(2024)Ramesh, Hu, Chaimalas, Mehta, Sessa, Bou~Ammar, and Bogunovic]{ramesh2024group}
Shyam~Sundhar Ramesh, Yifan Hu, Iason Chaimalas, Viraj Mehta, Pier~Giuseppe Sessa, Haitham Bou~Ammar, and Ilija Bogunovic.
\newblock Group robust preference optimization in reward-free rlhf.
\newblock \emph{Advances in Neural Information Processing Systems}, 37:\penalty0 37100--37137, 2024.

\bibitem[Setlur et~al.(2025)Setlur, Rajaraman, Levine, and Kumar]{setlur2025scaling}
Amrith Setlur, Nived Rajaraman, Sergey Levine, and Aviral Kumar.
\newblock Scaling test-time compute without verification or rl is suboptimal.
\newblock \emph{arXiv preprint arXiv:2502.12118}, 2025.

\bibitem[Shao et~al.(2024)Shao, Wang, Zhu, Xu, Song, Bi, Zhang, Zhang, Li, Wu, et~al.]{shao2024deepseekmath}
Zhihong Shao, Peiyi Wang, Qihao Zhu, Runxin Xu, Junxiao Song, Xiao Bi, Haowei Zhang, Mingchuan Zhang, YK~Li, Y~Wu, et~al.
\newblock Deepseekmath: Pushing the limits of mathematical reasoning in open language models.
\newblock \emph{arXiv preprint arXiv:2402.03300}, 2024.

\bibitem[Shen et~al.(2025)Shen, Liu, Li, Fang, Ma, Liao, Shen, Zhang, Zhao, Zhang, et~al.]{shen2025vlm}
Haozhan Shen, Peng Liu, Jingcheng Li, Chunxin Fang, Yibo Ma, Jiajia Liao, Qiaoli Shen, Zilun Zhang, Kangjia Zhao, Qianqian Zhang, et~al.
\newblock Vlm-r1: A stable and generalizable r1-style large vision-language model.
\newblock \emph{arXiv preprint arXiv:2504.07615}, 2025.

\bibitem[Singh et~al.(2023)Singh, Co-Reyes, Agarwal, Anand, Patil, Garcia, Liu, Harrison, Lee, Xu, et~al.]{singh2023beyond}
Avi Singh, John~D Co-Reyes, Rishabh Agarwal, Ankesh Anand, Piyush Patil, Xavier Garcia, Peter~J Liu, James Harrison, Jaehoon Lee, Kelvin Xu, et~al.
\newblock Beyond human data: Scaling self-training for problem-solving with language models.
\newblock \emph{arXiv preprint arXiv:2312.06585}, 2023.

\bibitem[Snell et~al.(2024)Snell, Lee, Xu, and Kumar]{snell2024scaling}
Charlie Snell, Jaehoon Lee, Kelvin Xu, and Aviral Kumar.
\newblock Scaling llm test-time compute optimally can be more effective than scaling model parameters.
\newblock \emph{arXiv preprint arXiv:2408.03314}, 2024.

\bibitem[Stiennon et~al.(2020)Stiennon, Ouyang, Wu, Ziegler, Lowe, Voss, Radford, Amodei, and Christiano]{stiennon2020learning}
Nisan Stiennon, Long Ouyang, Jeffrey Wu, Daniel Ziegler, Ryan Lowe, Chelsea Voss, Alec Radford, Dario Amodei, and Paul~F Christiano.
\newblock Learning to summarize with human feedback.
\newblock \emph{Advances in Neural Information Processing Systems}, 33:\penalty0 3008--3021, 2020.

\bibitem[Team et~al.(2025)Team, Du, Gao, Xing, Jiang, Chen, Li, Xiao, Du, Liao, et~al.]{team2025kimi}
Kimi Team, Angang Du, Bofei Gao, Bowei Xing, Changjiu Jiang, Cheng Chen, Cheng Li, Chenjun Xiao, Chenzhuang Du, Chonghua Liao, et~al.
\newblock Kimi k1. 5: Scaling reinforcement learning with llms.
\newblock \emph{arXiv preprint arXiv:2501.12599}, 2025.

\bibitem[Team(2025)]{qwq32b}
Qwen Team.
\newblock Qwq-32b: Embracing the power of reinforcement learning, March 2025.
\newblock URL \url{https://qwenlm.github.io/blog/qwq-32b/}.

\bibitem[Wang et~al.(2024{\natexlab{a}})Wang, Meng, Liang, and Zhou]{wang2024drt}
Jiaan Wang, Fandong Meng, Yunlong Liang, and Jie Zhou.
\newblock Drt-o1: Optimized deep reasoning translation via long chain-of-thought.
\newblock \emph{arXiv e-prints}, pp.\  arXiv--2412, 2024{\natexlab{a}}.

\bibitem[Wang et~al.(2024{\natexlab{b}})Wang, Fang, Wan, Wen, Zhu, Liu, Gong, Song, Chen, Ni, et~al.]{wang2024openr}
Jun Wang, Meng Fang, Ziyu Wan, Muning Wen, Jiachen Zhu, Anjie Liu, Ziqin Gong, Yan Song, Lei Chen, Lionel~M Ni, et~al.
\newblock Openr: An open source framework for advanced reasoning with large language models.
\newblock \emph{arXiv preprint arXiv:2410.09671}, 2024{\natexlab{b}}.

\bibitem[Wang \& Zhou(2024)Wang and Zhou]{wang2024chain}
Xuezhi Wang and Denny Zhou.
\newblock Chain-of-thought reasoning without prompting.
\newblock \emph{arXiv preprint arXiv:2402.10200}, 2024.

\bibitem[Wei et~al.(2022)Wei, Wang, Schuurmans, Bosma, Xia, Chi, Le, Zhou, et~al.]{wei2022chain}
Jason Wei, Xuezhi Wang, Dale Schuurmans, Maarten Bosma, Fei Xia, Ed~Chi, Quoc~V Le, Denny Zhou, et~al.
\newblock Chain-of-thought prompting elicits reasoning in large language models.
\newblock \emph{Advances in neural information processing systems}, 35:\penalty0 24824--24837, 2022.

\bibitem[Weng(2025)]{weng2025think}
Lilian Weng.
\newblock Why we think.
\newblock \emph{lilianweng.github.io}, May 2025.
\newblock URL \url{https://lilianweng.github.io/posts/2025-05-01-thinking/}.

\bibitem[Weston \& Sukhbaatar(2023)Weston and Sukhbaatar]{weston2023system}
Jason Weston and Sainbayar Sukhbaatar.
\newblock System 2 attention (is something you might need too).
\newblock \emph{arXiv preprint arXiv:2311.11829}, 2023.

\bibitem[Wu et~al.(2024)Wu, Sun, Li, Welleck, and Yang]{wu2024inference}
Yangzhen Wu, Zhiqing Sun, Shanda Li, Sean Welleck, and Yiming Yang.
\newblock Inference scaling laws: An empirical analysis of compute-optimal inference for problem-solving with language models.
\newblock \emph{arXiv preprint arXiv:2408.00724}, 2024.

\bibitem[{xAI}(2025)]{xai2025grok35}
{xAI}.
\newblock {Grok 3.5: Advanced reasoning AI model by xAI}.
\newblock \url{https://grok.x.ai/}, 2025.

\bibitem[Xu et~al.(2025)Xu, Hao, Zong, Wang, Zhang, Wang, Lan, Gong, Ouyang, Meng, et~al.]{xu2025towards}
Fengli Xu, Qianyue Hao, Zefang Zong, Jingwei Wang, Yunke Zhang, Jingyi Wang, Xiaochong Lan, Jiahui Gong, Tianjian Ouyang, Fanjin Meng, et~al.
\newblock Towards large reasoning models: A survey of reinforced reasoning with large language models.
\newblock \emph{arXiv preprint arXiv:2501.09686}, 2025.

\bibitem[Yang et~al.(2024)Yang, Yang, Zhang, Hui, Zheng, Yu, Li, Liu, Huang, Wei, et~al.]{yang2024qwen2}
An~Yang, Baosong Yang, Beichen Zhang, Binyuan Hui, Bo~Zheng, Bowen Yu, Chengyuan Li, Dayiheng Liu, Fei Huang, Haoran Wei, et~al.
\newblock Qwen2. 5 technical report.
\newblock \emph{arXiv preprint arXiv:2412.15115}, 2024.

\bibitem[Yang et~al.(2025)Yang, Ma, Lin, and Wei]{yang2025thinkingoptimalscalingtesttimecompute}
Wenkai Yang, Shuming Ma, Yankai Lin, and Furu Wei.
\newblock Towards thinking-optimal scaling of test-time compute for llm reasoning, 2025.
\newblock URL \url{https://arxiv.org/abs/2502.18080}.

\bibitem[Ying et~al.(2024)Ying, Zhang, Li, Zhou, Shao, Fei, Ma, Hong, Liu, Wang, et~al.]{ying2024internlm}
Huaiyuan Ying, Shuo Zhang, Linyang Li, Zhejian Zhou, Yunfan Shao, Zhaoye Fei, Yichuan Ma, Jiawei Hong, Kuikun Liu, Ziyi Wang, et~al.
\newblock Internlm-math: Open math large language models toward verifiable reasoning.
\newblock \emph{arXiv preprint arXiv:2402.06332}, 2024.

\bibitem[Yu et~al.(2024)Yu, Xu, Weston, and Kulikov]{yu2024distilling}
Ping Yu, Jing Xu, Jason Weston, and Ilia Kulikov.
\newblock Distilling system 2 into system 1.
\newblock \emph{arXiv preprint arXiv:2407.06023}, 2024.

\bibitem[Yu et~al.(2025)Yu, Zhang, Zhu, Yuan, Zuo, Yue, Fan, Liu, Liu, Liu, et~al.]{yu2025dapo}
Qiying Yu, Zheng Zhang, Ruofei Zhu, Yufeng Yuan, Xiaochen Zuo, Yu~Yue, Tiantian Fan, Gaohong Liu, Lingjun Liu, Xin Liu, et~al.
\newblock Dapo: An open-source llm reinforcement learning system at scale.
\newblock \emph{arXiv preprint arXiv:2503.14476}, 2025.

\bibitem[Yu et~al.(2023)Yu, Tao, Chen, Sun, and Yang]{yu2023mathcal}
Zishun Yu, Yunzhe Tao, Liyu Chen, Tao Sun, and Hongxia Yang.
\newblock $\mathcal{B}$-coder: Value-based deep reinforcement learning for program synthesis.
\newblock \emph{arXiv preprint arXiv:2310.03173}, 2023.

\bibitem[Yuan et~al.(2023)Yuan, Yuan, Li, Dong, Lu, Tan, Zhou, and Zhou]{yuan2023scaling}
Zheng Yuan, Hongyi Yuan, Chengpeng Li, Guanting Dong, Keming Lu, Chuanqi Tan, Chang Zhou, and Jingren Zhou.
\newblock Scaling relationship on learning mathematical reasoning with large language models.
\newblock \emph{arXiv preprint arXiv:2308.01825}, 2023.

\bibitem[Zelikman et~al.(2022)Zelikman, Wu, Mu, and Goodman]{zelikman2022star}
Eric Zelikman, Yuhuai Wu, Jesse Mu, and Noah Goodman.
\newblock Star: Bootstrapping reasoning with reasoning.
\newblock \emph{Advances in Neural Information Processing Systems}, 35:\penalty0 15476--15488, 2022.

\bibitem[Zelikman et~al.(2024)Zelikman, Harik, Shao, Jayasiri, Haber, and Goodman]{zelikman2024quiet}
Eric Zelikman, Georges Harik, Yijia Shao, Varuna Jayasiri, Nick Haber, and Noah~D Goodman.
\newblock Quiet-star: Language models can teach themselves to think before speaking.
\newblock \emph{arXiv preprint arXiv:2403.09629}, 2024.

\bibitem[Zhang et~al.(2024{\natexlab{a}})Zhang, Wu, Lei, Che, Li, Xie, Huang, Zhang, Pavone, Li, et~al.]{zhang2024llama}
Di~Zhang, Jianbo Wu, Jingdi Lei, Tong Che, Jiatong Li, Tong Xie, Xiaoshui Huang, Shufei Zhang, Marco Pavone, Yuqiang Li, et~al.
\newblock Llama-berry: Pairwise optimization for o1-like olympiad-level mathematical reasoning.
\newblock \emph{arXiv preprint arXiv:2410.02884}, 2024{\natexlab{a}}.

\bibitem[Zhang et~al.(2025{\natexlab{a}})Zhang, Lin, Hou, Feng, and Li]{zhang2025adaptthink}
Jiajie Zhang, Nianyi Lin, Lei Hou, Ling Feng, and Juanzi Li.
\newblock Adaptthink: Reasoning models can learn when to think.
\newblock \emph{arXiv preprint arXiv:2505.13417}, 2025{\natexlab{a}}.

\bibitem[Zhang et~al.(2025{\natexlab{b}})Zhang, Ruan, Ma, Zhu, Zhao, Li, Chen, Zeng, and Cai]{zhang2025continue}
Xiaoyun Zhang, Jingqing Ruan, Xing Ma, Yawen Zhu, Haodong Zhao, Hao Li, Jiansong Chen, Ke~Zeng, and Xunliang Cai.
\newblock When to continue thinking: Adaptive thinking mode switching for efficient reasoning.
\newblock \emph{arXiv preprint arXiv:2505.15400}, 2025{\natexlab{b}}.

\bibitem[Zhang et~al.(2024{\natexlab{b}})Zhang, Wu, Yang, Shu, Xiao, Kong, and Sang]{zhang2024o1}
Yuxiang Zhang, Shangxi Wu, Yuqi Yang, Jiangming Shu, Jinlin Xiao, Chao Kong, and Jitao Sang.
\newblock o1-coder: an o1 replication for coding.
\newblock \emph{arXiv preprint arXiv:2412.00154}, 2024{\natexlab{b}}.

\bibitem[Zhao et~al.(2024)Zhao, Yin, Zeng, Wang, Shi, Lyu, Wang, Luo, and Zhang]{zhao2024marco}
Yu~Zhao, Huifeng Yin, Bo~Zeng, Hao Wang, Tianqi Shi, Chenyang Lyu, Longyue Wang, Weihua Luo, and Kaifu Zhang.
\newblock Marco-o1: Towards open reasoning models for open-ended solutions.
\newblock \emph{arXiv preprint arXiv:2411.14405}, 2024.

\end{thebibliography}
\bibliographystyle{sample}

\newpage
\appendix

\section{Acknowledgments}

Chakraborty and Huang are supported by DARPA Transfer from Imprecise and Abstract Models to Autonomous Technologies (TIAMAT) 80321, DARPA HR001124S0029-AIQ-FP-019, DOD-AFOSR-Air Force Office of Scientific Research under award number FA9550-23-1-0048, National Science Foundation NSF-IIS-2147276 FAI, National Science Foundation NAIRR240045, National Science Foundation TRAILS Institute (2229885). Private support was provided by Peraton and Open Philanthropy.

\section{Broader Impact}
\label{app:broader_impact}

In this paper, we present a systematic study of test-time scaling, examining how prompting models to \textit{``Wait \& think''} at inference affects performance. Our experiments uncover a critical limitation: excessive sequential reasoning can degrade accuracy, whereas encouraging multiple parallel reasoning paths consistently yields better results. We hope these findings will inspire research on designing methods that guarantee reliable performance gains from test-time compute scaling. We do not anticipate any direct harmful consequences of our work.

\section{Additional Context of Related Works}
\label{sec:related_works2}
\textbf{LLMs for reasoning tasks.} In recent years, LLMs have advanced significantly, demonstrating impressive reasoning capabilities~\citep{lewkowycz2022solving, guo2025deepseek, openai2024gpt4, openo1, hou2025thinkprune, muennighoff2025s1, aggarwal2025l1}. Approaches to enhance the performance of LLMs on reasoning-intensive tasks~\citep{cobbe2021gsm8k, li2024gsm, AIME, hendrycks2021measuring, hendrycks2021measuringcode} typically fall into three categories: continued pretraining on extensive corpora~\citep{guo2025deepseek, shao2024deepseekmath, lewkowycz2022solving, azerbayev2023llemma, ying2024internlm, yang2024qwen2}, targeted fine-tuning for specialized reasoning skills~\citep{arora2025traininglanguagemodelsreason, aggarwal2025l1, singh2023beyond, shao2024deepseekmath, zelikman2022star, yuan2023scaling, ramesh2024group, yu2023mathcal, weston2023system, yu2024distilling, zelikman2024quiet, kumar2024training, hou2025thinkprune}, and employing additional computational resources during inference, known as test-time scaling~\citep{muennighoff2025s1, setlur2025scaling, snell2024scaling, wu2024inference, wang2024chain}. Out of these approaches, test-time scaling for reasoning tasks has recently gained significant attention, with studies~\citep{wei2022chain, snell2024scaling, setlur2025scaling, muennighoff2025s1} demonstrating substantial improvements in base model performance by allocating additional compute resources during inference or test-time. 

Concurrent with our study, \citet{marjanovic2025deepseek} analyzed the impact of reasoning trace length on model performance in the AIME dataset, likewise finding that increasing the number of thinking tokens generally leads to a decrease in average accuracy, a trend also observed in an extended analysis by \citep{weng2025think, muennighoff2025s1}. In this work, we extend these analyses across multiple datasets and a range of state-of-the-art reasoning models. Furthermore, we provide an intuitive explanation for this phenomenon and demonstrate that parallel thinking can offer a more effective strategy for utilizing test-time compute budgets.

\section{Software and Hardware}
\label{app:specs}

We run all experiments with Python 3.11.11, PyTorch 2.6.0, and Transformers 4.49.0. We run all experiments on Nvidia RTX A6000 GPUs.

\section{Additional Results on Test-time Budget Control}
\label{app:additional_results}

\subsection{Minimum Thinking setup}
\label{app:min_think}

\begin{figure}[!ht]
    \centering
    \includegraphics[width=\linewidth]{figures/minimum_thinking_tokens.pdf}
    \caption{\textbf{Results for \emph{Minimum thinking tokens} setup.} We visualize accuracy as a function of the thinking token budget for three benchmark datasets: GSM-8K~\citep{cobbe2021gsm8k}, MATH-500~\citep{lightman2023let}, and AIME~\citep{AIME}. We report results for three open-source reasoning models: DeepSeek-R1-Distill-Qwen-1.5B (1st row),  DeepSeek-R1-Distill-Qwen-7B (2nd row), and DeepSeek-R1-Distill-Llama-8B (3rd row)~\citep{guo2025deepseek}. For each prompt, we iteratively append the token \textit{``Wait''} to the reasoning trace until the thinking trajectory includes at least $t_{\min}$ tokens of thought. We observe a decline in accuracy with an increase in the thinking token budget.}
    \label{fig:min_ttbc}
\end{figure}

\textbf{TTBC 3: Minimum thinking tokens.} For this approach, we enforce a lower bound $t_{\min}$ on the number of thinking tokens per prompt. We suppress the predicted end-of-thinking delimiter and iteratively append the token \textit{``Wait''} to the reasoning trace until the cumulative count of thinking tokens reaches $t_{\min}$. Once this threshold is met, we explicitly append the end-of-thinking delimiter (\texttt{</think>}) to transition the model out of the thinking phase and prompt it to provide its final response. Fixing $t_{\min}$, we ensure that every thinking trajectory includes at least $t_{\min}$ tokens of thought. For this analysis, we vary $t_{\min}\in\{256,512,1024,2048,4096,8192,16384\}$ to study the scaling effects of different minimum token budgets. 

\noindent\textbf{Observations.} Figure~\ref{fig:min_ttbc} illustrates the trade-off between minimum thinking tokens and average accuracy under the \emph{Minimum Thinking} TTBC. We report results for three open-source models—DeepSeek-R1-Distill-Qwen-1.5B, DeepSeek-R1-Distill-Qwen-7B, and DeepSeek-R1-Distill-Llama-8B—across three mathematical benchmarks: GSM-8K, MATH-500, and AIME 2024. Consistent with the trends observed in the \emph{Wait \& Think More} (Figure 2, main paper) and \emph{Exact Thinking} (Figure 3, main paper) TTBCs, we find a non-monotonic relationship between accuracy and the minimum thinking token threshold across all models and datasets. These results further reinforce that increasing the computation budget does not necessarily lead to better reasoning outcomes, highlighting the need to rethink prevailing test-time reasoning strategies and move beyond the simplistic assumption that more computation inherently improves performance.

\subsection{Extended analysis on role of variance in test-time scaling}
\label{app:extended_variance_analysis}

In Section 3.1 of the main paper, our investigation on GSM-8K and MATH-500 using DeepSeek-R1-Distill-Qwen-1.5B revealed that the initial accuracy gains from test-time scaling, up to a certain average number of thinking tokens, are largely a mirage, driven more by randomness than genuine model refinement. Here, we extend this analysis to additional state-of-the-art open-source reasoning models: DeepSeek-R1-Distill-Qwen-1.5B and DeepSeek-R1-Distill-Llama-8B. Results for DeepSeek-R1-Distill-Qwen-1.5B are shown in Figure~\ref{fig:entropy_qwen7b}, and for DeepSeek-R1-Distill-Llama-8B in Figure~\ref{fig:entropy_llama}. Across all models, we observe a consistent trend: the entropy of the policy distribution increases with the length of thinking tokens. This observation further reinforces our main takeaway that the \emph{``Wait \& Think More''} approach is not an optimal way to allocate the test-time thinking budget.

\begin{figure}[!ht]
    \centering
    \includegraphics[width=0.75\linewidth]{figures/qwen7b_entropy_y.pdf}
    \caption{We investigate how extended thinking impacts policy uncertainty by visualizing entropy as a function of additional thinking tokens on GSM-8K~\citep{cobbe2021gsm8k} and MATH-500~\citep{hendrycks2021measuring}. We find that increasing the number of thinking tokens initially raises entropy and leads to better performance. However, after a certain threshold, further increases in thinking tokens cause a sharp rise in entropy, which negatively affects performance. For our experiments, we use the ``Wait \& Think more'' setup with DeepSeek-R1-Distill-Qwen-7B.}
    \label{fig:entropy_qwen7b}
\end{figure}

\begin{figure}[!ht]
    \centering
    \includegraphics[width=0.75\linewidth]{figures/llama_entropy_y.pdf}
    \caption{We investigate how extended thinking impacts policy uncertainty by visualizing entropy as a function of additional thinking tokens on GSM-8K~\citep{cobbe2021gsm8k} and MATH-500~\citep{hendrycks2021measuring}. We find that increasing the number of thinking tokens initially raises entropy and leads to better performance. However, after a certain threshold, further increases in thinking tokens cause a sharp rise in entropy, which negatively affects performance. For our experiments, we use the ``Wait \& Think more'' setup, and the model is DeepSeek-R1-Distill-Llama-8B.}
    \label{fig:entropy_llama}
\end{figure}

\section{Generation Examples}

In the figures below, we visualize example generations for three test-time budget control setups: \emph{Wait \& Think More}, \emph{Exact Thinking Token}, and \emph{Minimum Thinking Token}. In all these examples, extended thinking leads to an incorrect answer.

\begin{tcolorbox}[floatplacement=h,float, width=\columnwidth, colback=
mygreen!1!,
colframe=
gray, title=Generation with \textit{``Wait \& Think More''}]\label{gen:wait} 
    \textbf{Prompt:} Josh decides to try flipping a house.  He buys a house for \$80,000 and then puts in \$50,000 in repairs.  This increased the value of the house by 150\%.  How much profit did he make?. 

\textcolor{gray}{\noindent\rule{\textwidth}{1pt}}
    \textbf{Standard Thinking:} Okay, so Josh bought a house for \$80,000. Then he spent \$50,000 on repairs. After that, the value of the house increased by 150\%. I need to figure out how much profit he made from this entire process. First, let me break down the problem step by step. Josh starts with a house worth \$80,000. Then he adds \$50,000 to it for repairs. So, the total cost he incurred is the initial price plus the repair 
cost. That would be \$80,000 + \$50,000 = \$130,000. Hmm, wait, is that right? Or is the repair cost an additional investment, so maybe it\'s just \$50,000 on top of the \$80,000? I think it\'s the latter. So, the total cost is \$80,000 + \$50,000 = \$130,000. Now, $\ldots\ldots$ it is the new value minus the total cost, which is \$200,000 - \$130,000 = \$70,000. Yes, that seems consistent. So, I think the profit is
 \$70,000 \verb|\n|\textcolor{darkgreen}{\textbf{Final Answer} Josh made a profit of \boxed{70000} dollars.}

\textcolor{gray}{\noindent\rule{\textwidth}{1pt}}
\textit{\textbf{Wait \& Think more} (5k thinking tokens)}: Okay, so Josh bought a house for \$80,000. Then he spent \$50,000 on repairs. After that, the value of the house increased by 150\%. I need to figure out how much profit he made from this entire process. First, let me break down the problem step by step $\ldots\ldots\ldots$ \verb|\n|Wait, I think I'm making a mistake. Let me just do it step by step. 1. Initial cost: \$80,000 2. Repair cost: \$50,000 3. Total cost: \$80,000 + \$50,000 = \$130,000 4. Value increase: 150\% of \$80,000 = \$120,000 5. New value: \$80,000 + \$120,000 = \$200,000 6. Profit: \$200,000 - \$130,000 = \$70,000 So, the profit is
\$70,000. Wait, no, hold on. \verb|\n|Wait, I think I'm making a mistake here. The increase is 150\% of the original value, which is \$80,000, so \$120,000. So, the new value is \$200,000. Profit is \$200,000 - \$130,000 = \$70,000. Yes, that's correct. So, the profit is \$70,000 \verb|\n|\textcolor{darkgreen}{\textbf{Final Answer} \boxed{70000}}
 
\textcolor{gray}{\noindent\rule{\textwidth}{1pt}}
\textit{\textbf{Wait \& Think more} (12k thinking tokens)}:  Okay, so Josh bought a house for \$80,000. Then he spent \$50,000 on repairs. After that, the value of the house increased by 150\%. I need to figure out how much profit he made from this entire process. First, let me break down the problem step by step $\ldots\ldots\ldots\ldots$ \verb|\n|Wait, no, hold on. Wait, I think I'm overcomplicating this. Let me just write down the numbers clearly - Initial cost: \$80,000 - Repair cost: \$50,000 $\ldots$ so the total profit is -10000. \verb|\n|\textcolor{red}{\textbf{Final Answer} \boxed{-10000}}

\end{tcolorbox}

\begin{tcolorbox}[floatplacement=h,float, width=\columnwidth, colback=
mygreen!1!,
colframe=
gray, title=Generation with \textit{Exact Thinking} TTBC]\label{gen:exact}

\textbf{Prompt:} Simplify \text{$\sqrt{242}$}

\textcolor{gray}{\noindent\rule{\textwidth}{1pt}} 

\textit{\textbf{Exact Thinking} (256 thinking tokens):} To simplify the square root of 242, I first need to factorize 242 into its prime components. I start by dividing 242 by 2, which gives me 121. Next, I recognize that 121 is a perfect square because 11 multiplied by 11 equals 121. So, the prime factorization of 242 is 2 multiplied by 11 squared. Taking the square root of this, I can separate the square root of 11 squared from the square root of 2. The square root of 11 squared is 11, and the square root of 2 remains as is. Therefore, the simplified form of the square root of 242 is 11 times the square root of 2. Wait, that doesn't seem right. Let me double-check my steps. I think I made a mistake in the prime factorization. Let me try again. 242 divided by 2 is 121, which is correct. But 121 is 11 squared, so the prime factors are 2 and 11 squared. So, $\sqrt{242} = \sqrt{2 * 11^2} = 11 * \sqrt{2}$. Yes, that makes sense. So the simplified form is 11 times the square root of 2 $\ldots \ldots$ Then, $\sqrt{242} = \sqrt{121 * 2} = \sqrt{121} * \sqrt{2} = 11 * \sqrt{2}$. Yes, that confirms it. So the simplified form is $11 \sqrt{2}$. \verb|\n|\textcolor{darkgreen}{\textbf{Final Answer}: \boxed{11 \sqrt{2}}}

\textcolor{gray}{\noindent\rule{\textwidth}{1pt}}
    \textit{\textbf{Exact Thinking} (2048 thinking tokens):} To simplify the square root of 242, I first need to factorize 242 into its prime components. I start by dividing 242 by 2, which gives me 121. Next, I recognize that 121 is a perfect square because 11 multiplied by 11 equals 121. So, the prime factorization of 242 is 2 multiplied by 11 squared. Taking the square root of this, I can separate the square root of 11 squared from the square root of 2. The square root of 11 squared is 11, and the square root of 2 remains as is. Therefore, the simplified form of the square root of 242 is 11 times the square root of 2. Wait, that doesn't seem right. $\ldots \ldots \ldots$ Wait, but I'm still confused because I thought maybe I made a mistake in the prime factorization. Let me try factoring 242 again. 242 divided by 2 is 121. 121 is 11 * 11. So, 242 = 2 * \verb|\n|\textcolor{red}{\textbf{Final Answer}: \boxed{2}}

\end{tcolorbox}

\begin{tcolorbox}[floatplacement=h,float, width=\columnwidth, colback=
mygreen!1!,
colframe=
gray, title=Generation with \textit{Minimum Thinking} TTBC]\label{gen:min}

\textbf{Prompt:} The Doubtfire sisters are driving home with 7 kittens adopted from the local animal shelter when their mother calls to inform them that their two house cats have just had kittens. She says that Patchy, the first cat, has had thrice the number of adopted kittens, while Trixie, the other cat, has had 12. How many kittens does the Doubtfire family now have?

\textcolor{gray}{\noindent\rule{\textwidth}{1pt}}
    \textit{\textbf{Minimum Thinking} (256 thinking tokens):} Patchy has had three times the number of adopted kittens. Since there are 7 kittens, Patchy has had 3 * 7 = 21 kittens. Trixie has had 12 kittens. To find the total number of kittens the family now has, I add the kittens from both cats: 21 (from Patchy) + 12 (from Trixie) = 33 kittens. Therefore, the Doubtfire family now has 33 kittens in total. Wait, that doesn\'t seem right. The initial 7 kittens are adopted, and then the cats have had more kittens. So the total number of kittens would be the original 7 plus the kittens Patchy and Trixie have had. So, Patchy has 21 kittens, Trixie has 12, and they already have 7. Adding those together: 21 + 12 + 7 = 40 kittens in total. \verb|\n|Wait, I think I made a mistake earlier. The initial 7 kittens are adopted, and then the cats have had more kittens. So the total number of kittens is the original 7 plus the kittens Patchy and Trixie have had. Patchy has 3 times the adopted kittens: 3 * 7 = 21.Trixie has 12 kittens. So, total kittens = 7 (adopted) + 21 (Patchy) + 12 (Trixie) = 40 kittens. Yes, that makes sense now. \verb|\n|Wait, no, that can't be right. The original 7 kittens are already part of the family, and the cats have had more kittens. So the total number of kittens is the original 7 plus the kittens Patchy and Trixie have had. So, Patchy has 21, Trixie has 12, and they already have 7. Total kittens = 7 + 21 + 12 = 40. \verb|\n|\textcolor{darkgreen}{\textbf{Final Answer:} \boxed{40}.}

\textcolor{gray}{\noindent\rule{\textwidth}{1pt}}
    \textit{\textbf{Minimum Thinking} (2048 thinking tokens):} First, I need to understand the information given. The Doubtfire sisters have 7 kittens that they adopted from a local shelter. Their mother informs them that their two house cats, Patchy and Trixie, have had kittens. Patchy has had three times the number of adopted kittens. Since there are 7 kittens, Patchy has had 3 * 7 = 21 kittens. Trixie has had 12 kittens. To find the total number of kittens the family now has, I add the kittens from both cats: 21 (from Patchy) + 12 (from Trixie) = 33 kittens. Therefore, the Doubtfire family now has 33 kittens in total. \verb|\n|Wait, that doesn't seem right. The initial 7 kittens are adopted, and then the cats have had more kittens. So the total number of kittens would be the original 7 plus the kittens Patchy and Trixie have had. So, Patchy has 21 kittens, Trixie has 12, and they already have 7. Adding those together: 21 + 12 + 7 = 40 kittens in total. \verb|\n|Wait, I think I made a mistake earlier. The initial 7 kittens are adopted, and then the cats have had more kittens. So the total number of kittens is the original 7 plus the kittens Patchy and Trixie have had. Patchy has 3 times the adopted kittens: 3 * 7 = 21. $\ldots \ldots \ldots$ I think I've spent enough time on this. The answer is 33. \verb|\n|\textcolor{red}{\textbf{Final Answer:} \boxed{33}}.

\end{tcolorbox}

\end{document}